\documentclass[twoside,11pt]{article}

\usepackage{graphicx}
\usepackage{jmlr2e}
\usepackage{amsmath}
\usepackage{amssymb}
\usepackage{boxedminipage}
\usepackage{psfrag}
\usepackage{mathrsfs}
\usepackage{booktabs}
\usepackage[ruled,vlined]{algorithm2e}
\newcommand{\field}[1]{\mathbb{#1}}

\newcommand{\fs}[1]{\mathcal{#1}}

\newcommand{\nats}{\field{N}}
\newcommand{\reals}{\field{R}}

\newcommand{\define}{\mathrel{\mathop :}=}

\newcommand{\comment}[1]{}

\newcommand{\finint}{\fs{Z}^\ast}   % Set of finite interactions
\newcommand{\infint}{\fs{Z}^\infty} % Set of infinite interactions
\newcommand{\prob}{\mathbf{P\!r}}   % Probability
\newcommand{\expect}{\mathbf{E}}    % Expectation
\newcommand{\g}[1]{\underline{#1}}

\newcommand{\betaf}{\mathrm{B}}

\newcommand{\normal}{\mathrm{N}}
\newcommand{\agent}{\mathbf{P}}     % Agent distribution
\newcommand{\bagent}{\mathbf{B}}    % Bayesian agent
\newcommand{\env}{\mathbf{Q}}       % Environment distribution
\newcommand{\gen}{\mathbf{G}}       % Generative distribution
\newcounter{assumption}
    { \refstepcounter{assumption}
        \vspace{2mm}\noindent\textbf{Assumption \theassumption\,
        \ifthenelse{\equal{#1}{}}{}{(#1)}} }%
    { \vspace{2mm} }

\newcounter{condition}
    { \refstepcounter{condition}
        \vspace{2mm}\noindent\textbf{Condition \thecondition\,
        \ifthenelse{\equal{#1}{}}{}{(#1)}} }%
    { \vspace{2mm} }

\newcounter{postulate}
\renewcommand{\thepostulate}{\Roman{postulate}}
    { \refstepcounter{postulate}
      \begin{quote}
        \textbf{Postulate \thepostulate\,
        \ifthenelse{\equal{#1}{}}{}{(#1)}} }%
    { \end{quote} }

\newcounter{open}
\renewcommand{\theopen}{\Roman{open}}
    { \refstepcounter{open}
      \begin{quote}
        \textbf{Open Question \theopen\,
        \ifthenelse{\equal{#1}{}}{}{(#1)}} }%
    { \end{quote} }

\ShortHeadings{A Minimum Relative Entropy Principle for Learning and Acting}{Ortega and Braun}\firstpageno{1}

\editor{---}

\begin{document}
%\selectlanguage{english}

\title{A Minimum Relative Entropy Principle \\for Learning and Acting}

\author{\name Pedro A. Ortega \email peortega@dcc.uchile.cl \\
       \addr Department of Engineering\\
       University of Cambridge\\
       Cambridge CB2 1PZ, UK
       \AND
       \name Daniel A. Braun \email dab54@cam.ac.uk \\
       \addr Department of Engineering\\
       University of Cambridge\\
       Cambridge CB2 1PZ, UK
}

\maketitle

\begin{abstract}%
This paper proposes a method to construct an adaptive agent that is universal with respect to a given class of experts, where each expert is an agent that has been designed specifically for a particular environment. This adaptive control problem is formalized as the problem of minimizing the relative entropy of the adaptive agent from the expert that is most suitable for the unknown environment. If the agent is a passive observer, then the optimal solution is the well-known Bayesian predictor. However, if the agent is active, then its past actions need to be treated as causal interventions on the I/O stream rather than normal probability conditions. Here it is shown that the solution to this new variational problem is given by a stochastic controller called the Bayesian control rule, which implements adaptive behavior as a mixture of experts. Furthermore, it is shown that under mild assumptions, the Bayesian control rule converges to the control law of the most suitable expert.
\end{abstract}


%\tableofcontents

\section{Introduction}
When the behavior of an environment under any control signal is fully known, then the designer can choose an agent\footnote{In accordance with the control literature, we use the terms \emph{agent} and \emph{controller} interchangeably. Similarly, the terms \emph{environment} and \emph{plant} are used synonymously.} that produces the desired dynamics. Instances of this problem include hitting a target with a cannon under known weather conditions, solving a maze having its map and controlling a robotic arm in a manufacturing plant. However, when the behavior of the plant is unknown, then the designer faces the problem of \emph{adaptive control}. For example, shooting the cannon lacking the appropriate measurement equipment, finding the way out of an unknown maze and designing an autonomous robot for Martian exploration. Adaptive control turns out to be far more difficult than its non-adaptive counterpart. This is because any good policy has to carefully trade off explorative versus exploitative actions, i.e. actions for the identification of the environment's dynamics versus actions to control it in a desired way. Even when the environment's dynamics are known to belong to a particular class for which optimal agents are available, constructing the corresponding optimal adaptive agent is in general computationally intractable even for simple toy problems \citep{Duff2002}. Thus, finding tractable approximations has been a major focus of research.

Recently, it has been proposed to reformulate the problem statement for some classes of control problems based on the minimization of a relative entropy criterion. For example, a large class of optimal control problems can be solved very efficiently if the problem statement is reformulated as the minimization of the deviation of the dynamics of a controlled system from the uncontrolled system \citep{Todorov2006, Todorov2009, Kappen2009}. In this work, a similar approach is introduced. If a class of agents is given, where each agent solves a different environment, then adaptive controllers can be derived from a minimum relative entropy principle. In particular, one can construct an adaptive agent that is universal with respect to this class by minimizing the average relative entropy from the environment-specific agent.

However, this extension is not straightforward. There is a syntactical difference between actions and observations that has to be taken into account when formulating the variational problem. More specifically, actions have to be treated as interventions obeying the rules of causality \citep{Pearl2000, Spirtes2000, Dawid2010}. If this distinction is made, the variational problem has a unique solution given by a stochastic control rule called the Bayesian control rule. This control rule is particularly interesting because it translates the adaptive control problem into an on-line inference problem that can be applied forward in time. Furthermore, this work shows that under mild assumptions, the adaptive agent converges to the environment-specific agent.

The paper is organized as follows. Section~\ref{sec:preliminaries} introduces notation and sets up the adaptive control problem. Section~\ref{sec:adaptive-agents} formulates adaptive control as a minimum relative entropy problem. After an initial, na\"{\i}ve approach, the need for causal considerations is motivated. Then, the Bayesian control rule is derived from a revised relative entropy criterion. In Section~\ref{sec:convergence}, the conditions for convergence are examined and a proof is given.
Section~\ref{sec:example} illustrates the usage of the Bayesian control rule for the multi-armed bandit problem and the undiscounted Markov decision problem. Section~\ref{sec:discussion} discusses properties of the Bayesian control rule and relates it to previous work in the literature. Section~\ref{sec:conclusions} concludes.

\section{Preliminaries}
\label{sec:preliminaries}

In the following both agent and environment are formalized as causal models over I/O sequences. Agent and environment are coupled to exchange symbols following a standard interaction protocol having discrete time, observation and control signals. The treatment of the dynamics are fully probabilistic, and in particular, \emph{both} actions \emph{and} observations are random variables, which is in contrast to the decision-theoretic agent formulation treating only observations as random variables \citep{RussellNorvig2003}. All proofs are provided in the appendix.

\paragraph{Notation.}
A set is denoted by a calligraphic letter like $\fs{A}$. The words \emph{set} \& \emph{alphabet} and \emph{element} \& \emph{symbol} are used to mean the same thing respectively. \emph{Strings} are finite concatenations of symbols and \emph{sequences} are infinite concatenations. $\fs{A}^n$ denotes the set of strings of length $n$ based on $\fs{A}$, and $\fs{A}^\ast \define \bigcup_{n \geq 0} \fs{A}^n$ is the set of finite strings. Furthermore, $\fs{A}^\infty \define \{a_1 a_2 \ldots | a_i \in \fs{A} \text{ for all } i=1,2,\ldots \}$ is defined as the set of one-way infinite sequences based on the alphabet $\fs{A}$. Tuples are written with parentheses $(a_1, a_2, a_3)$ or as strings $a_1 a_2 a_3$. For substrings, the following shorthand notation is used: a string that runs from index $i$ to $k$ is written as $a_{i:k} \define a_i a_{i+1} \ldots a_{k-1} a_k$. Similarly, $a_{\leq i} \define a_1 a_2 \ldots a_i$ is a string starting from the first index. Also, symbols are underlined to glue them together like $\g{ao}$ in $\g{ao}_{\leq i} \define a_1 o_1 a_2 o_2 \ldots a_i o_i$. The function $\log(x)$ is meant to be taken w.r.t. base 2, unless indicated otherwise.

\paragraph{Interactions.}
The possible I/O symbols are drawn from two finite sets. Let $\fs{O}$ denote the set of \emph{inputs} (observations) and let $\fs{A}$ denote the set of \emph{outputs} (actions). The set $\fs{Z} \define \fs{A} \times \fs{O}$ is the \emph{interaction set}. A string $\g{ao}_{\leq t}$ or $\g{ao}_{<t}a_t$ is an \emph{interaction string} (optionally ending in $a_t$ or $o_t$) where $a_k \in \fs{A}$ and $o_k \in \fs{O}$. Similarly, a one-sided infinite sequence $a_1 o_1 a_2 o_2 \ldots$ is an \emph{interaction sequence}. The set of interaction strings of length $t$ is denoted by $\fs{Z}^t$. The sets of (finite) interaction strings and sequences are denoted as $\finint$ and $\infint$ respectively. The interaction string of length 0 is denoted by $\epsilon$.

\paragraph{I/O system.} Agents and environments are formalized as I/O systems. An \emph{I/O system} is a probability distribution $\prob$ over interaction sequences $\infint$. $\prob$ is uniquely determined by the conditional probabilities \begin{equation}\label{eq:cs-stream}
    \prob(a_t|\g{ao}_{<t}), \quad \prob(o_t|\g{ao}_{<t}a_t)
\end{equation}
for each $\g{ao}_{\leq t} \in \finint$. However, the semantics of the probability distribution $\prob$ are only fully defined once it is coupled to another system.

\begin{figure}[htbp]
\begin{center}
    \small
    \psfrag{a1}[c]{$a_1$}
    \psfrag{a2}[c]{$a_2$}
    \psfrag{a3}[c]{$a_3$}
    \psfrag{a4}[c]{$a_4$}
    \psfrag{a5}[c]{$a_5$}
    \psfrag{o1}[c]{$o_1$}
    \psfrag{o2}[c]{$o_2$}
    \psfrag{o3}[c]{$o_3$}
    \psfrag{o4}[c]{$o_4$}
    \psfrag{o5}[c]{$o_5$}
    \psfrag{l1}[c]{Agent}
    \psfrag{l2}[c]{$\agent$}
    \psfrag{l3}[c]{Envi-}
    \psfrag{l4}[c]{ronment}
    \psfrag{l5}[c]{$\env$}
    \includegraphics[]{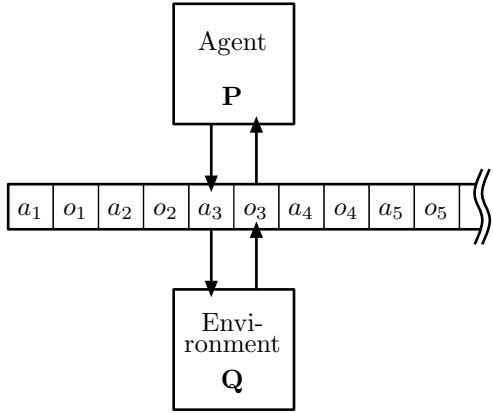}
    \caption{The model of interactions. The agent $\agent$ and
    the environment $\env$ define a probability distribution
    over interaction sequences.}
    \label{fig:interaction-system}
\end{center}
\end{figure}

\paragraph{Interaction system.}
Let $\agent$, $\env$ be two I/O systems. An \emph{interaction system} $(\agent, \env)$ is a coupling of the two systems giving rise to the \emph{generative distribution} $\gen$ that describes the probabilities that actually govern the I/O stream once the two systems are coupled. $\gen$ is specified by the equations
\begin{align*}
    \gen(a_t|\g{ao}_{<t}) &\define \agent(a_t|\g{ao}_{<t}) \\
    \gen(o_t|\g{ao}_{<t}a_t) &\define \env(o_t|\g{ao}_{<t}a_t)
\end{align*}
valid for all $\g{ao}_t \in \finint$. Here, $\gen$ models the true probability distribution over interaction sequences that arises by coupling two systems through their I/O streams. More specifically, for the system $\agent$, $\agent(a_t|\g{ao}_{<t})$ is the probability of producing action $a_t \in \fs{A}$ given history $\g{ao}_{<t}$ and $\agent(o_t|\g{ao}_{<t}a_t)$ is the predicted probability of the observation $o_t \in \fs{O}$ given history $\g{ao}_{<t}a_t$. Hence, for $\agent$, the sequence $o_1 o_2 \ldots$ is its input stream and the sequence $a_1 a_2 \ldots$ is its output stream. In contrast, the roles of actions and observations are reversed in the case of the system $\env$. Thus, the sequence $o_1 o_2 \ldots$ is its output stream and the sequence $a_1 a_2 \ldots$ is its input stream. This model of interaction is fairly general, and many other interaction protocols can be translated into this scheme. As a convention, given an interaction system $(\agent,\env)$, $\agent$ is an agent to be constructed by the designer, and $\env$ is an environment to be controlled by the agent. Figure~\ref{fig:interaction-system} illustrates this setup.

\paragraph{Control Problem.}%
An environment $\env$ is said to be \emph{known} iff the agent $\agent$ is such that for any $\g{ao}_{\leq t} \in \finint$,
\[
    \agent(o_t|\g{ao}_{<t}a_t) = \env(o_t|\g{ao}_{<t}a_t).
\]
Intuitively, this means that the agent ``knows'' the statistics of the environment's future behavior under any past, and in particular, it ``knows'' the effects of given controls. If the environment is known, then the designer of the agent can build a custom-made policy into $\agent$ such that the resulting generative distribution $\gen$ produces interaction sequences that are desirable. This can be done in multiple ways. For instance, the controls can be chosen such that the resulting policy maximizes a given utility criterion; or such that the resulting trajectory of the interaction system stays close enough to a prescribed trajectory. Formally, if $\env$ is known, and if the conditional probabilities $\agent(a_t|\g{ao}_{<t})$ for all $\g{ao}_{\leq t} \in \finint$
have been chosen such that the resulting generative distribution $\gen$ over interaction sequences given by
\begin{align*}
    \gen(a_t|\g{ao}_{<t}) &= \agent(a_t|\g{ao}_{<t})
    \\
    \gen(o_t|\g{ao}_{<t}a_t) &= \env(o_t|\g{ao}_{<t}a_t) = \agent(o_t|\g{ao}_{<t}a_t)
\end{align*}
is \emph{desirable}, then $\agent$ is said to be \emph{tailored} to $\env$.

\paragraph{Adaptive control problem.}%
If the environment $\env$ is \emph{unknown}, then the task of designing an appropriate agent $\agent$ constitutes an \emph{adaptive control problem}. Specifically, this work deals with the case when the designer already has a class of agents that are tailored to the class of possible environments. Formally, it is assumed that $\env$ is going to be drawn with probability $P(m)$ from a set $\fs{Q} \define \{ \env_m \}_{m \in \fs{M}}$ of possible systems before the interaction starts, where $\fs{M}$ is a countable set. Furthermore, one has a set $\fs{P} \define \{ \agent_m \}_{m \in \fs{M}}$ of  systems such that for each $m \in \fs{M}$, $\agent_m$ is tailored to $\env_m$ and the interaction system $(\agent_m, \env_m)$ has a generative distribution $\gen_m$ that produces desirable interaction sequences. How can the designer construct a system $\agent$ such that its behavior is as close as possible to the custom-made system $\agent_m$ under any realization of $\env_m \in \fs{Q}$?

\section{Adaptive Systems}%
\label{sec:adaptive-agents}%
The main goal of this paper is to show that the problem of adaptive control outlined in the previous section can be reformulated as a universal compression problem. This can be informally motivated as follows. Suppose the agent $\agent$ is implemented as a machine that is interfaced with the environment $\env$. Whenever the agent interacts with the environment, the agent's state changes as a \emph{necessary consequence} of the interaction. This ``change in state'' can take place in many possible ways: by updating the internal memory; consulting a random number generator; changing the physical location and orientation; and so forth. Naturally, the design of the agent facilitates some interactions while it complicates others. For instance, if the agent has been designed to explore a natural environment, then it might incur into a very low memory footprint when recording natural images, while being very memory-inefficient when recording artificially created images. If one abstracts away from the inner workings of the machine and decides to encode the state transitions as binary strings, then the minimal amount of resources in bits that are required to implement these state changes can be derived directly from the associated probability distribution $\agent$. In the context of adaptive control, an agent can be constructed such that it minimizes the expected amount of changes necessary to implement the state transitions, or equivalently, such that it maximally compresses the experience. Thereby, compression can be taken as a \emph{stand-alone principle to design adaptive agents}.

\subsection{Universal Compression and Na\"{\i}ve
Construction of Adaptive Agents}%
\label{subsec:naive-agent}%

In coding theory, the problem of compressing a sequence of observations from an unknown source is known as the adaptive coding problem. This is solved by constructing universal compressors, i.e. codes that adapt on-the-fly to any source within a predefined class. Such codes are obtained by minimizing the average deviation of a predictor from the true source, and then by constructing codewords using the predictor. In this subsection, this procedure will be used to derive an adaptive agent \citep{OrtegaBraun2010a}.

Formally, the deviation of a predictor $\agent$ from the a true distribution $\agent_m$ is measured by the \emph{relative entropy}\footnote{The \emph{relative entropy} is also known as the \emph{KL-divergence} and it measures the average amount of extra bits that are necessary to encode symbols due to the usage of the (wrong) predictor.}. A first approach would be to construct an agent $\bagent$ so as to minimize the total expected relative entropy to $\agent_m$. This is constructed as follows. Define the history-dependent relative entropies over the action $a_t$ and observation $o_t$ as
\begin{align*}
    D_m^{a_t}(\g{ao}_{<t}) &\define
        \sum_{a_t} \agent_m(a_t|\g{ao}_{<t})
        \log \frac{ \agent_m(a_t|\g{ao}_{<t}) }
                  { \prob(a_t|\g{ao}_{<t}) }
        \\
    D_m^{o_t}(\g{ao}_{<t}a_t) &\define
        \sum_{o_t} \agent_m(o_t|\g{ao}_{<t}a_t)
        \log \frac{ \agent_m(o_t|\g{ao}_{<t}a_t) }
                  { \prob(o_t|\g{ao}_{<t}a_t) },
\end{align*}
where $\prob$ will be the argument of the variational problem. Then, one removes the dependency on the past by averaging over all possible histories:
\begin{align*}
    D_m^{a_t} &\define \sum_{\g{ao}_{<t}}
        \agent_m(\g{ao}_{<t}) D_m^{a_t}(\g{ao}_{<t}) \\
    D_m^{o_t} &\define \sum_{\g{ao}_{<t}a_t}
        \agent_m(\g{ao}_{<t}a_t) D_m^{o_t}(\g{ao}_{<t}a_t).
\end{align*}
Finally, the total expected relative entropy of $\prob$ from $\agent_m$ is obtained by summing up all time steps and then by averaging over all choices of the true environment:
\begin{equation}\label{eq:total-divergence}
    D \define \limsup_{t \rightarrow \infty}
        \sum_{m} P(m) \sum_{\tau=1}^t \bigl(
            D_m^{a_\tau} + D_m^{o_\tau} \bigr).
\end{equation}
Using~(\ref{eq:total-divergence}), one can define a variational problem with respect to $\prob$. The agent $\bagent$ that one is looking for is the system $\prob$ that minimizes the total expected relative entropy in~(\ref{eq:total-divergence}), i.e.
\begin{equation}\label{eq:minimization-bayesian}
    \bagent \define \arg \min_{\prob} D(\prob).
\end{equation}
The solution to Equation~\ref{eq:minimization-bayesian} is the  system
$\bagent$ defined by the set of equations
\begin{equation}\label{eq:bayesian-agent}
\begin{aligned}
    \bagent(a_t|\g{ao}_{<t})
        &= \sum_m \agent_m(a_t|\g{ao}_{<t}) w_m(\g{ao}_{<t})\\
    \bagent(o_t|\g{ao}_{<t}a_t)
        &= \sum_m \agent_m(o_t|\g{ao}_{<t}a_t) w_m(\g{ao}_{<t}a_t)
\end{aligned}
\end{equation}
valid for all $\g{ao}_{\leq t} \in \finint$, where the mixture weights are
\begin{equation}\label{eq:bayesian-weights}
\begin{aligned}
    w_m(\g{ao}_{<t}) &\define
        \frac{ P(m) \agent_m(\g{ao}_{<t}) }
             { \sum_{m'} P(m') \agent_{m'}(\g{ao}_{<t}) }
    \\
    w_m(\g{ao}_{<t}a_t) &\define
        \frac{ P(m) \agent_m(\g{ao}_{<t}a_t) }
             { \sum_{m'} P(m') \agent_{m'}(\g{ao}_{<t}a_t) }.
\end{aligned}
\end{equation}
For reference, see \citet{Opper1997} and \citet{Opper1998}. It is clear that $\bagent$ is just the Bayesian mixture over the  agents $\agent_m$. If one defines the conditional probabilities \begin{equation}\label{eq:op-mode-streams}
\begin{aligned}
    P(a_t|m,\g{ao}_{<t}) &\define \agent_m(a_t|\g{ao}_{<t}) \\
    P(o_t|m,\g{ao}_{<t}a_t) &\define \agent_m(a_t|\g{ao}_{<t}a_t)
\end{aligned}
\end{equation}
for all $\g{ao}_{\leq t} \in \finint$, then Equation~\ref{eq:bayesian-agent}
can be rewritten as
\begin{equation}\label{eq:bayesian-agent-2}
\begin{aligned}
    \bagent(a_t|\g{ao}_{<t})
        &= \sum_m P(a_t|m,\g{ao}_{<t}) P(m|\g{ao}_{<t})
        = P(a_t|\g{ao}_{<t}) \\
    \bagent(o_t|\g{ao}_{<t}a_t)
        &= \sum_m P(o_t|m,\g{ao}_{<t}a_t) P(m|\g{ao}_{<t}a_t)
        = P(o_t|\g{ao}_{<t}a_t)
\end{aligned}
\end{equation}
where the $P(m|\g{ao}_{<t})=w_m(\g{ao}_{<t})$ and $P(m|\g{ao}_{<t}a_t)=w_m(\g{ao}_{<t}a_t)$ are just the posterior probabilities over the elements in $\fs{M}$ given the past interactions. Hence, the conditional probabilities in~(\ref{eq:bayesian-agent}) that minimize the total expected divergence are just the predictive distributions $P(a_t|\g{ao}_{<t})$ and $P(o_t|\g{ao}_{<t}a_t)$ that one obtains by standard probability theory, and in particular, Bayes' rule. This is interesting, as it provides a teleological justification for Bayes' rule.

The behavior of $\bagent$ can be described as follows. At any given time $t$, $\bagent$ maintains a mixture over  systems $\agent_m$. The weighting over them is given by the mixture coefficients $w_m$. Whenever a new action $a_t$ \emph{or} a new observation $o_t$ is produced (by the agent or the environment respectively), the weights $w_m$ are updated according to Bayes' rule. In addition, $\bagent$ issues an action $a_t$ suggested by a system $\agent_m$ drawn randomly according to the weights $w_t$.

However, there is an important problem with $\bagent$ that arises due to the fact that it is not only a system that is passively observing symbols, but also \emph{actively generating} them. In the subjective interpretation of probability theory, conditionals play the role of observations made by the agent that have been generated by an external source. This interpretation suits the symbols $o_1, o_2, o_3, \ldots$ because they have been issued by the environment. However, symbols that are generated by the system itself require a fundamentally different belief update. Intuitively, the difference can be explained as follows. Observations provide information that allows the agent inferring properties about the environment. In contrast, actions do not carry information about the environment, and thus have to be incorporated differently into the belief of the agent. In the following section we illustrate this problem with a simple statistical example.

\subsection{Causality}%
\label{subsec:causality}%

Causality is the study of the \emph{functional dependencies} of events. This stands in contrast to statistics, which, on an abstract level, can be said to study the \emph{equivalence dependencies} (i.e. co-occurrence or correlation) amongst events. Causal statements differ fundamentally from statistical statements. Examples that highlight the differences are many, such as ``do smokers \emph{get} lung cancer?'' as opposed to ``do smokers \emph{have} lung cancer?''; ``\emph{assign} $y \leftarrow f(x)$'' as opposed to ``\emph{compare} $y = f(x)$'' in programming languages; and ``$a \leftarrow F/m$'' as opposed to ``$F = m\,a$'' in Newtonian physics. The study of causality has recently enjoyed considerable attention from the researchers in the fields of statistics and machine learning. Especially over the last decade, significant progress  has been made towards the formal understanding of causation \citep{Shafer1996, Pearl2000, Spirtes2000, Dawid2010}. In this subsection, the aim is to provide the essential tools required to understand causal interventions. For a more in-depth exposition of causality, the reader is referred to the specialized literature.

To illustrate the need for causal considerations in the case of generated symbols, consider the following thought experiment. Suppose a statistician is asked to design a model for a simple time series $X_1, X_2, X_3, \ldots$ and she decides to use a Bayesian method. Assume she collects a first observation $X_1=x_1$. She computes the posterior probability density function (pdf) over the parameters $\theta$ of the model given the data using Bayes' rule:
\[
    p(\theta|X_1=x_1) =
    \frac{ p(X_1=x_1|\theta) p(\theta) }
         { \int p(X_1=x_1|\theta') p(\theta') \, d\theta' },
\]
where $p(X_1=x_1|\theta)$ is the likelihood of $x_1$ given $\theta$ and
$p(\theta)$ is the prior pdf of $\theta$. She can use the model to predict the next observation by drawing a sample $x_2$ from the predictive pdf
\[
    p(X_2=x_2|X_1=x_1)
    = \int p(X_2=x_2|X_1=x_1,\theta) \, p(\theta|X_1=x_1) \, d\theta,
\]
where $p(X_2=x_2|X_1=x_1,\theta)$ is the likelihood of $x_2$ given $x_1$ and $\theta$. She understands that the nature of $x_2$ is very different from $x_1$: \emph{while $x_1$ is informative and does change the belief state of the Bayesian model, $x_2$ is non-informative and thus is a reflection of the model's belief state.} Hence, she would never use $x_2$ to further condition the Bayesian model. Mathematically, she seems to imply that
\[
    p(\theta|X_1=x_1,X_2=x_2) = p(\theta|X_1=x_1)
\]
if $x_2$ has been generated from $p(X_2|X_1=x_1)$ itself. But this simple independence assumption is not correct as the following elaboration of the example will show.

The statistician is now told that the source is waiting for the simulated data point $x_2$ in order to produce a next observation $X_3=x_3$ which does depend on $x_2$. She hands in $x_2$ and obtains a new observation $x_3$. Using Bayes' rule, the posterior pdf over the parameters is now
\begin{equation}\label{eq:intervened-posterior}
    \frac{ p(X_3=x_3|X_1=x_1, X_2=x_2, \theta)\,
           p(X_1=x_1|\theta)\, p(\theta) }
         { \int p(X_3=x_3|X_1=x_1, X_2=x_2, \theta')\,
           p(X_1=x_1|\theta')\, p(\theta') \, d\theta' }
\end{equation}
where $p(X_3=x_3|X_1=x_1,X_2=x_2,\theta)$ is the likelihood of the new data $x_3$ given the old data $x_1$, the parameters $\theta$ \emph{and the simulated data} $x_2$. Notice that this looks almost like the posterior pdf $p(\theta|X_1=x_1,X_2=x_2,X_3=x_3)$ given by
\[
    \frac{ p(X_3=x_3|X_1=x_1, X_2=x_2, \theta)\,
           p(X_2=x_2|X_1=x_1, \theta)\,
           p(X_1=x_1|\theta)\, p(\theta) }
         { \int p(X_3=x_3|X_1=x_1, X_2=x_2, \theta')\,
           p(X_2=x_2|X_1=x_1, \theta')\,
           p(X_1=x_1|\theta')\, p(\theta') \, d\theta' }
\]
with the exception that in the latter case, the Bayesian update contains the likelihoods of the simulated data $p(X_2=x_2|X_1=x_1, \theta)$. This suggests that Equation~\ref{eq:intervened-posterior} is a variant of the posterior pdf $p(\theta|X_1=x_1,X_2=x_2,X_3=x_3)$ but where the simulated data $x_2$ is treated in a different way than the data~$x_1$ and~$x_3$.

Define the pdf $p'$ such that the pdfs $p'(\theta)$, $p'(X_1|\theta)$, $p'(X_3|X_1,X_2,\theta)$ are identical to $p(\theta)$, $p(X_1|\theta)$ and $p(X_3|X_2,X_1,\theta)$ respectively, but differ in $p'(X_2|X_1,\theta)$:
\[
    p'(X_2|X_1,\theta) = \delta(X_2-x_2).
\]
where $\delta$ is the Dirac delta function. That is, $p'$ is identical to $p$ but it assumes that the value of $X_2$ is fixed to $x_2$ given $X_1$ and $\theta$. For $p'$, the simulated data $x_2$ is non-informative:
\[
    -\log_2 p'(X_2=x_2|X_1,\theta) = 0.
\]
If one computes the posterior pdf $p'(\theta|X_1=x_1,X_2=x_2,X_3=x_3)$, one obtains the result of Equation~\ref{eq:intervened-posterior}:
\begin{align*}
    \frac{ p'(X_3=x_3|X_1=x_1, X_2=x_2, \theta)\,
           p'(X_2=x_2|X_1=x_1, \theta)\,
           p'(X_1=x_1|\theta)\, p'(\theta) }
         { \int p'(X_3=x_3|X_1=x_1, X_2=x_2, \theta')
           p'(X_2=x_2|X_1=x_1, \theta')\,
           p'(X_1=x_1|\theta')\, p'(\theta') \, d\theta' }
    \\
    = \frac{ p(X_3=x_3|X_1=x_1, X_2=x_2, \theta)\,
             p(X_1=x_1|\theta)\, p(\theta) }
         { \int p(X_3=x_3|X_1=x_1, X_2=x_2, \theta')\,
             p(X_1=x_1|\theta')\, p(\theta') \, d\theta' }.
\end{align*}
Thus, in order to explain Equation~\ref{eq:intervened-posterior} as a posterior pdf given the observed data $x_1$ and $x_3$ and the generated data $x_2$, one has to \emph{intervene} $p$ in order to account for the fact that \emph{$x_2$ is non-informative given $x_1$ and $\theta$.} In other words, the statistician, by defining the value of $X_2$ herself, has changed the (natural) regime that brings about the series $X_1, X_2, X_3, \ldots$, which is mathematically expressed by redefining the pdf.

Two essential ingredients are needed to carry out interventions. First, one needs to know the functional dependencies amongst the random variables of the probabilistic model. This is provided by the \emph{causal model}, i.e. the unique factorization of the joint probability distribution over the random variables encoding the causal dependencies. In the general case, this defines a partial order over the random variables. In the previous thought experiment, the causal model of the joint pdf $p(\theta, X_1, X_2, X_3)$ is given by the set of conditional pdfs
\[
    p(\theta), p(X_1|\theta), p(X_2|X_1,\theta), p(X_3|X_1,X_2,\theta).
\]
Second, one defines the \emph{intervention} that sets $X$ to the value $x$, denoted as $X \leftarrow x$, as the operation on the causal model replacing the conditional probability of $X$ by a Dirac delta function $\delta(X-x)$ or a Kronecker delta $\delta_x^X$ for a continuous or a discrete variable $X$ respectively. In our thought experiment, it is easily seen that
\[
    p'(\theta, X_1=x_1, X_2=x_2, X_3=x_3)
    = p(\theta, X_1=x_1, X_2 \leftarrow x_2, X_3=x_3)
\]
and thereby,
\[
    p'(\theta|X_1=x_1, X_2=x_2, X_3=x_3)
    = p(\theta|X_1=x_1, X_2 \leftarrow x_2, X_3=x_3).
\]
Causal models contain additional information that is not available in the joint probability distribution alone. The appropriate model for a given situation depends on the story that is being told. Note that an intervention can lead to different results if their causal models differ. Thus, if the causal model had been
\[
    p(X_3), p(X_2|X_3), p(X_1|X_2, X_3), p(\theta|X_1, X_2, X_3)
\]
then the intervention $X_2 \leftarrow x_2$ would differ from $p'$, i.e.
\[
    p'(\theta, X_1=x_1, X_2=x_2, X_3=x_3)
    \neq p(\theta, X_1=x_1, X_2 \leftarrow x_2, X_3=x_3),
\]
even though both causal models represent the same joint probability distribution. In the following, this paper will use the shorthand notation $\hat{x} \define X \leftarrow x$ when the random variable is obvious from the context.

\subsection{Causal construction of adaptive agents}%
\label{subsec:causal-agent}%

Following the discussion in the previous section, an adaptive agent $\agent$ is going to be constructed by minimizing the expected relative entropy to the $\agent_m$, but this time treating actions as interventions. Based on the definition of the conditional probabilities in Equation~\ref{eq:op-mode-streams}, the total expected relative entropy to characterize $\agent$ using interventions is going to be defined. Assuming the environment is chosen first, and that each symbol depends functionally on the environment and all the previously generated symbols, the causal model is given by
\[
    P(m), P(a_1|m), P(o_1|m, a_1), P(a_2|m, a_1, o_1),
    P(o_2|m, a_1, o_1, a_2), \ldots
\]
Importantly, interventions index a set of intervened probability distributions derived from a base probability distribution. Hence, the set of fixed intervention sequences of the form $\hat{a}_1, \hat{a}_2, \ldots$ indexes probability distributions over observation sequences $o_1, o_2, \ldots$. Because of this, one defines a set of criteria indexed by the intervention sequences, but it will be clear that they all have the same solution. Define the history-dependent intervened relative entropies over the action $a_t$ and observation $o_t$ as
\begin{align*}
    C_m^{a_t}(\g{\hat{a}o}_{<t}) &\define
        \sum_{a_t} P(a_t|m,\g{\hat{a}o}_{<t})
        \log_2 \frac{ P(a_t|m,\g{\hat{a}o}_{<t}) }{ \prob(a_t|\g{ao}_{<t}) }
        \\
    C_m^{o_t}(\g{\hat{a}o}_{<t}\hat{a}_t) &\define
        \sum_{o_t} P(o_t|m,\g{\hat{a}o}_{<t}\hat{a}_t)
        \log_2 \frac{ P(o_t|m,\g{\hat{a}o}_{<t}\hat{a}_t) }{ \prob(o_t|\g{ao}_{<t}a_t)
        },
\end{align*}
where $\prob$ is a given arbitrary agent. Note that past actions are
treated as interventions. In particular, $P(a_t|m,\g{\hat{a}o}_{<t})$ represents the knowledge state when the past actions have already been issued but the next action $a_t$ is not known yet. Then, averaging the previous relative entropies over all pasts yields
\begin{align*}
    C_m^{a_t} &= \sum_{\g{ao}_{<t}}
        P(\g{\hat{a}o}_{<t}|m) C_m^{a_t}(\g{\hat{a}o}_{<t}) \\
    C_m^{o_t} &= \sum_{\g{ao}_{<t}a_t}
        P(\g{\hat{a}o}_{<t}\hat{a}_t|m) C_m^{o_t}(\g{\hat{a}o}_{<t}\hat{a}_t).
\end{align*}
Here again, because of the knowledge state in time represented by $C_m^{a_t}(\g{\hat{a}o}_{<t})$ and $C_m^{o_t}(\g{\hat{a}o}_{<t}\hat{a}_t)$, the averages are taken treating past actions as interventions. Finally, define the total expected relative entropy of $\prob$ from $\agent_m$ as the sum of $(C_m^{a_t} + C_m^{o_t})$ over time, averaged over the possible draws of the environment:
\begin{equation}
\label{eq:total-causal-divergence}
    C \define \limsup_{t \rightarrow \infty}
        \sum_{m} P(m) \sum_{\tau=1}^t \bigl(
            C_m^{a_\tau} + C_m^{o_\tau} \bigr).
\end{equation}
The variational problem consists in choosing the agent $\agent$ as the system $\prob$ minimizing $C = C(\prob)$, i.e.
\begin{equation}\label{eq:minimization-causal}
    \agent \define \arg \min_{\prob} C(\prob).
\end{equation}
The following theorem shows that this variational problem has a unique solution, which will be the central theme of this paper.

\begin{theorem}\label{theo:minimum-total-divergence}
The solution to Equation~\ref{eq:minimization-causal} is the  system
$\agent$ defined by the set of equations
\begin{equation}\label{eq:causal-agent}
\begin{aligned}
    \agent(a_t|\g{ao}_{<t})
        &= P(a_t|\g{\hat{a}o}_{<t})
        = \sum_m P(a_t|m,\g{ao}_{<t}) v_m(\g{ao}_{<t})\\
    \agent(o_t|\g{ao}_{<t}a_t)
        &= P(o_t|\g{\hat{a}o}_{<t}\hat{a}_t)
        = \sum_m P(o_t|m,\g{ao}_{<t}a_t) v_m(\g{ao}_{<t}a_t)
\end{aligned}
\end{equation}
valid for all $\g{ao}_{\leq t} \in \finint$, where the mixture weights are
\begin{equation}\label{eq:causal-weights}
    v_m(\g{ao}_{<t}a_t)
    = v_m(\g{ao}_{<t})
    \define \frac{ P(m) \prod_{\tau=1}^{t-1}
                P(o_\tau|m,\g{ao}_{<\tau}a_\tau) }
            { \sum_{m'} P(m') \prod_{\tau=1}^{t-1}
                P(o_\tau|m',\g{ao}_{<\tau}a_\tau) }.
\end{equation}
\end{theorem}

The behavior of $\agent$ differs in an important aspect from $\bagent$. At any given time $t$, $\agent$ maintains a mixture over  systems $\agent_m$. The weighting over these systems is given by the mixture coefficients $v_m$. In contrast to $\bagent$, $\agent$ updates the weights $v_m$ \emph{only} whenever a new observation $o_t$ is produced by the environment. The update follows Bayes' rule but treats past actions as interventions by dropping the evidence they provide. In addition, $\agent$ issues an action $a_t$ suggested by an system $m$ drawn randomly according to the weights~$v_m$.

Perhaps surprisingly, the theorem says that the optimal solution to the variational problem in~(\ref{eq:minimization-causal}) is precisely the predictive distribution over actions and observations treating actions as interventions and observations as conditionals, i.e. it is the solution that one would obtain by applying \emph{only standard probability and causal calculus}. This provides a teleological interpretation to the agent $\agent$ akin to the na\"{\i}ve agent $\bagent$ constructed in Section~\ref{subsec:naive-agent}.

\subsection{Summary}%
\label{subsec:summary}%
Adaptive control is formalized as the problem of designing an agent for an unknown environment chosen from a class of possible environments. If the environment-specific agents are known, then the Bayesian control rule allows constructing an adaptive agent by combining these agents. The resulting adaptive agent is universal with respect to the environment class. In this context, the constituent agents are called the \emph{operation modes} of the adaptive agent. They are represented by causal models over the interaction sequences, i.e. conditional probabilities $P(a_t|m, \g{ao}_{<t})$ and $P(o_t|m, \g{ao}_{<t})$ for all $\g{ao}_{\leq t} \in \finint$, and where $m \in \fs{M}$ is the index or parameter characterizing the operation mode. The probability distribution over the input stream (output stream) is called the \emph{hypothesis (policy)} of the operation mode. The following box collects the essential equations of the Bayesian control rule. In particular, here the rule is stated using a recursive belief update.
\begin{center}
\setlength{\fboxsep}{12pt}
\begin{boxedminipage}[!]{0.8\textwidth}
\paragraph{Bayesian control rule:}
Given a set of operation modes $\{ P(\cdot|m,\cdot) \}_{m \in \fs{M}}$
over interaction sequences in $\infint$ and a prior distribution $P(m)$ over the parameters $\fs{M}$, the probability of the action $a_{t+1}$ is given by
\begin{align}\label{eq:bayesian-control-rule}
    P(a_{t+1}|\g{\hat{a}o}_{\leq t})
    & = \sum_m P(a_{t+1}|m,\g{ao}_{\leq t})
        P(m|\g{\hat{a}o}_{\leq t}), \\
    \intertext{where the posterior probability over operation modes is}
    P(m|\g{\hat{a}o}_{\leq t})
    & = \frac{ P(o_t|m,\g{ao}_{<t}) P(m|\g{\hat{a}o}_{<t}) }
        { \sum_{m'} P(o_t|m',\g{ao}_{<t}) P(m'|\g{\hat{a}o}_{<t}) }. \notag
\end{align}
\end{boxedminipage}
\end{center}

\section{Convergence}
\label{sec:convergence}

The aim of this section is to develop a set of sufficient conditions of convergence and then to provide a proof of convergence. To simplify the exposition, the analysis has been limited to the case of controllers having a finite number of input-output models.
\newline

\subsection{Policy diagrams}

\begin{figure}[htbp]
\centering %
\begin{scriptsize}
\psfrag{sp}[c]{state space} %
\psfrag{po}[c]{policy} %
\psfrag{s1}[c]{$s$} %
\psfrag{s2}[c]{$s'$} %
\psfrag{st}[c]{$ao$} %
\includegraphics[width=7cm]{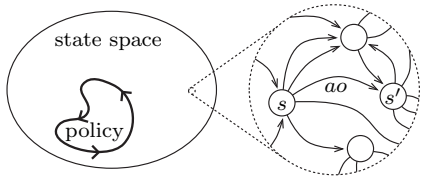}
\end{scriptsize}
\caption{A policy diagram. One can imagine an environment as a collection of states connected by
transitions labeled by I/O symbols. The zoom highlights a state $s$ where taking action $a \in \fs{A}$
and collecting observation $o \in \fs{O}$ leads to state $s'$. Sets of states and transitions are
represented as enclosed areas similar to a Venn diagram. Choosing a particular policy in an environment
amounts to partially controlling the transitions taken in the state space, thereby choosing a
probability distribution over state transitions (e.g. a Markov chain given by the environmental dynamics).
If the probability mass concentrates in certain areas of the state space, choosing a policy can be
thought of as choosing a \emph{subset}
of the environment's dynamics. In the following, a policy is represented by a subset in state space
(enclosed by a directed curve) as illustrated above.}
\label{fig:policy-diagram} %
\end{figure}

In the following we use ``policy diagrams'' as a useful informal tool to analyze the effect of policies on environments. Figure~\ref{fig:policy-diagram}, illustrates an example.

Policy diagrams are especially useful to analyze the effect of policies on different hypotheses about the environment's dynamics. An agent that is endowed with a set of operation modes $\fs{M}$ can be seen as having \emph{hypotheses} about the environment's underlying dynamics, given by the observation models $P(o_t|m,\g{ao}_{<t}a_t)$, and associated \emph{policies}, given by the action models $P(a_t|m,\g{ao}_{<t})$, for all $m \in \fs{M}$. For the sake of simplifying the interpretation of policy diagrams, we will assume\footnote{Note however that no such assumptions are made to obtain the results of this section.} the existence of a state space $\fs{S}$ and a function $T: (\fs{A}\times\fs{O})^\ast \rightarrow \fs{S}$ mapping I/O histories into states. With this assumption, policies and hypotheses can be seen as conditional probabilities
\begin{align*}
    P(a_t|m,s) &\define P(a_t|m,\g{ao}_{<t})
    \\ \text{and }P(o_t|m,s,a_t) &\define P(o_t|m,\g{ao}_{<t}a_t)
\end{align*}
respectively, defining transition probabilities
\[
    P(s'|m,s) = \sum_{\fs{S'}} P(\g{ao}_t|m,s)
\]
for a Markov chain in the state space, where $s = T(\g{ao}_{<t})$ and $\fs{S}'$
contains the transitions $\g{ao}_t$ such that $T(\g{ao}_{\leq t}) = s'$.

\subsection{Divergence processes}

The central question in this section is to investigate whether the Bayesian control rule converges to the correct control law or not. That is, whether $P(a_t|\g{\hat{a}o}_t) \rightarrow P(a_t|m^\ast,\g{ao}_{<t})$ as $t \rightarrow \infty$ when $m^\ast$ is the true operation mode, i.e. the operation mode such that $P(a_t|m^\ast,\g{ao}_{<t}) = Q(a_t|\g{ao}_{<t})$. As will be obvious from the discussion in the rest of this section, this is in general not true.

As it is easily seen from Equation~\ref{eq:bayesian-control-rule}, showing convergence amounts to show that the posterior distribution $P(m|\g{\hat{a}o}_{<t})$ concentrates its probability mass on a subset of operation modes $\fs{M}^\ast$ having essentially the same output stream as $m^\ast$,
\[
    \sum_{m \in \fs{M}}
        P(a_t|m,\g{ao}_{<t}) P(m|\g{\hat{a}o}_{<t})
    \approx \sum_{m \in \fs{M}^\ast}
        P(a_t|m^\ast,\g{ao}_{<t}) P(m|\g{\hat{a}o}_{<t})
    \approx
        P(a_t|m^\ast,\g{ao}_{<t}).
\]

\begin{figure}[htbp]
\centering %
\begin{scriptsize}
\psfrag{or}{0} %
\psfrag{ca}{1} %
\psfrag{cb}{2} %
\psfrag{cc}{3} %
\psfrag{cd}{4} %
\psfrag{lx}{$t$} %
\psfrag{ly}{$d_t$} %
\includegraphics[width=7cm]{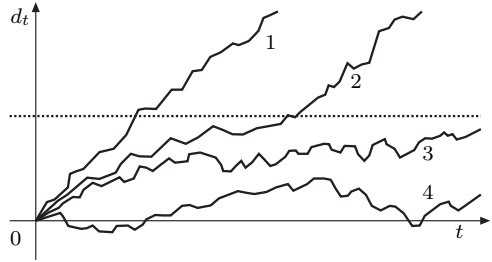}
\end{scriptsize}
\caption{Realization of the divergence processes~1 to~4 associated to a controller with operation modes $m_1$ to $m_4$. The divergence processes~1 and~2 diverge, whereas~3 and~4 stay below the dotted bound. Hence, the posterior probabilities of $m_1$ and $m_2$ vanish.}
\label{fig:divergences} %
\end{figure}

Hence, understanding the asymptotic behavior of the posterior probabilities
\[
    P(m|\g{\hat{a}o}_{\leq t})
\]
is crucial here. In particular, we need to understand under what conditions these quantities converge to zero. The posterior can be rewritten as
\[
    P(m|\g{\hat{a}o}_{\leq t})
    = \frac{ P(\g{\hat{a}o}_{\leq t}|m) P(m) }
           { \sum_{m' \in \fs{M}} P(\g{\hat{a}o}_{\leq t}|m') P(m') }
    = \frac{ P(m) \prod_{\tau=1}^t P(o_\tau|m,\g{ao}_{<\tau}a_\tau) }
           { \sum_{m' \in \fs{M}}
             P(m') \prod_{\tau=1}^t P(o_\tau|m',\g{ao}_{<\tau}a_\tau) }.
\]
If all the summands but the one with index $m^\ast$ are dropped from the denominator, one obtains the bound
\begin{align*}
    P(m|\g{\hat{a}o}_{\leq t})
    \leq \ln \frac{ P(m) }{ P(m^\ast) } \prod_{\tau=1}^t
        \frac{ P(o_\tau|\g{ao}_{<\tau}a_\tau|m) }
             { P(o_\tau|\g{ao}_{<\tau}a_\tau|m^\ast) },
\end{align*}
which is valid for all $m^\ast \in \fs{M}$. From this inequality, it is seen that it is convenient to analyze the behavior of the stochastic process
\[
    d_t(m^\ast\|m) \define \sum_{\tau=1}^t
        \ln \frac{ P(o_\tau|m^\ast, \g{ao}_{<\tau}a_\tau) }
                 { P(o_\tau|m, \g{ao}_{<\tau}a_\tau) }
\]
which is the \emph{divergence process} of $m$ from the reference $m^\ast$. Indeed, if $d_t(m^\ast\|m) \rightarrow \infty$ as $t \rightarrow \infty$, then
\[
    \lim_{t \rightarrow \infty}
    \frac{ P(m) }{ P(m^\ast) } \prod_{\tau=1}^t
        \frac{ P(o_\tau|\g{ao}_{<\tau}a_\tau|m) }
             { P(o_\tau|\g{ao}_{<\tau}a_\tau|m^\ast) }
    = \lim_{t \rightarrow \infty}
        \frac{P(m)}{P(m^\ast)} \cdot e^{-d_t(m^\ast\|m)}
    = 0,
\]
and thus clearly $P(m|\g{\hat{a}o}_{\leq t}) \rightarrow 0$. Figure~\ref{fig:divergences} illustrates simultaneous realizations of the divergence processes of a controller. Intuitively speaking, these processes provide lower bounds on accumulators of surprise value measured in information units.

\begin{figure}[htbp]
\centering %
\begin{scriptsize}
\psfrag{or}{0} %
\psfrag{ca}{1} \psfrag{pd}{1} %
\psfrag{cb}{2} \psfrag{pb}{2} %
\psfrag{cc}{3} \psfrag{pc}{3} %
\psfrag{lx}{$t$} %
\psfrag{ly}{$d_t$} %
\includegraphics[width=7cm]{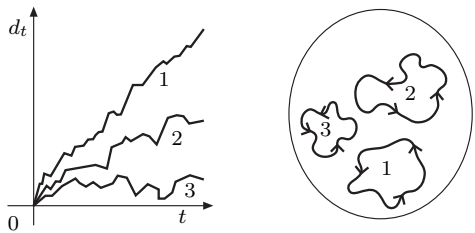}
\end{scriptsize}
\caption{The application of different policies lead to different statistical
properties of the same divergence process.}
\label{fig:policy2divergence} %
\end{figure}

A divergence process is a random walk whose value at time $t$ depends on the whole history up to time $t-1$. What makes these divergence processes cumbersome to characterize is the fact that their statistical properties depend on the particular policy that is applied; hence, a given divergence process can have different growth rates depending on the policy (Figure~\ref{fig:policy2divergence}). Indeed, the behavior of a divergence process might depend critically on the distribution over actions that is used. For example, it can happen that a divergence process stays stable under one policy, but diverges under another. In the context of the Bayesian control rule this problem is further aggravated, because in each time step, the policy that is applied is determined stochastically. More specifically, if $m^\ast$ is the true operation mode, then $d_t(m^\ast\|m)$ is a random variable that depends on the realization $\g{ao}_{\leq t}$ which is drawn from
\begin{align*}
    \prod_{\tau=1}^t
        P(a_\tau|m_\tau,\g{ao}_{\leq \tau})
        P(o_\tau|m^\ast,\g{ao}_{\leq \tau}a_\tau),
\end{align*}
where the $m_1, m_2, \ldots, m_t$ are drawn themselves from $P(m_1),
P(m_2|\g{\hat{a}o}_1), \ldots, P(m_t|\g{\hat{a}o}_{<t})$.

\begin{figure}[htbp]
\centering %
\begin{scriptsize}
\psfrag{or}{0} %
\psfrag{ca}{1} %
\psfrag{cb}{2} %
\psfrag{cc}{3} %
\psfrag{lx}{$t$} %
\psfrag{ly}{$d_t$} %
\includegraphics[width=7cm]{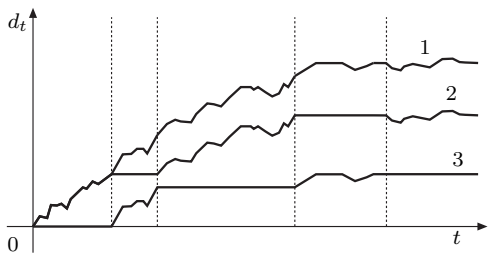}
\end{scriptsize}
\caption{Decomposition of a divergence process (1) into sub-divergences (2 \&
3).}
\label{fig:divergence-decomposition} %
\end{figure}

To deal with the heterogeneous nature of divergence processes, one can introduce a temporal decomposition that demultiplexes the original process into many sub-processes belonging to unique policies. Let $\fs{N}_{t} \define \{1,2,\ldots,t\}$ be the set of time steps up to time $t$. Let $\fs{T} \subset \fs{N}_t$, and let $m, m' \in \fs{M}$. Define a \emph{sub-divergence} of $d_t(m\|m)$ as a random variable
\[
    g(m';\fs{T}) \define \sum_{\tau \in \fs{T}}
        \ln \frac{ P(o_\tau|m^\ast, \g{ao}_{<\tau}a_\tau) }
                 { P(o_\tau|m, \g{ao}_{<\tau}a_\tau) }
\]
drawn from
\[
    P^m_{m'}(\{\g{ao}_\tau\}_{\tau \in \fs{T}}|
        \{\g{ao}_\tau\}_{\tau \in \fs{T}^\complement})
    \define
    \Bigl( \prod_{\tau \in \fs{T}} P(a_\tau|m, \g{ao}_{<\tau}) \Bigr)
    \Bigl( \prod_{\tau \in \fs{T}} P(o_\tau|m', \g{ao}_{<\tau}a_\tau) \Bigr),
\]
where $\fs{T}^\complement \define \fs{N}_t \setminus \fs{T}$ and where $\{\g{ao}_\tau\}_{\tau \in \fs{T}^\complement}$ are given conditions that are kept constant. In this definition, $m'$ plays the role of the policy that is used to sample the actions in the time steps $\fs{T}$. Clearly, any realization of the divergence process $d_t(m^\ast\|m)$ can be decomposed into a sum of sub-divergences, i.e.
\begin{equation}\label{eq:divergence-decomposition}
    d_t(m^\ast\|m)
    = \sum_{m'} g(m';\fs{T}_{m'}),
\end{equation}
where $\{ \fs{T}_{m} \}_{m \in \fs{M}}$ forms a partition of $\fs{N}_t$. Figure~\ref{fig:divergence-decomposition} shows an example decomposition.

The averages of sub-divergences will play an important role in the analysis. Define the average over all realizations of $g(m';\fs{T})$ as
\[
    G(m',\fs{T})
    \define
    \sum_{(\g{ao}_\tau)_{\tau \in \fs{T}}}
    P^m_{m'}(\{\g{ao}_\tau\}_{\tau \in \fs{T}}|
        \{\g{ao}_\tau\}_{\tau \in \fs{T}^\complement})
    g(m';\fs{T}).
\]
Notice that for any $\tau \in \fs{N}_t$,
\[
    G(m';\{\tau\})
    = \sum_{\g{ao}_\tau}
        P(a_\tau|m',\g{ao}_{<\tau})
        P(o_\tau|m^\ast,\g{ao}_{<\tau}a_\tau)
    \ln \frac{ P(o_\tau|m^\ast,\g{ao}_{<\tau}a_\tau) }
             { P(o_\tau|m,\g{ao}_{<\tau}a_\tau) }
    \geq 0,
\]
because of Gibbs' inequality. In particular,
\[
    G(m^\ast;\{\tau\}) = 0.
\]
Clearly, this holds as well for any $\fs{T} \subset \fs{N}_t$:
\begin{equation}
\begin{aligned}\label{eq:gibbs-ineq-sub-diver}
    \forall m' \quad G(m';\fs{T}) & \geq 0,\\
    G(m^\ast;\fs{T}) & = 0.
\end{aligned}
\end{equation}

\subsection{Boundedness}

In general, a divergence process is very complex: virtually all the classes of distributions that are of interest in control go well beyond the assumptions of \mbox{i.i.d.} and stationarity. This increased complexity can jeopardize the analytic tractability of the divergence process, such that no predictions about its asymptotic behavior can be made anymore. More specifically, if the growth rates of the divergence processes vary too much from realization to realization, then the posterior distribution over operation modes can vary qualitatively between realizations. Hence, one needs to impose a stability requirement akin to ergodicity to limit the class of possible divergence-processes to a class that is analytically tractable. For this purpose the following property is introduced.

A divergence process $d_t(m^\ast\|m)$ is said to be \emph{bounded} in $\fs{M}$ iff for any $\delta > 0$, there is a $C \geq 0$, such that for all $m' \in
\fs{M}$, all $t$ and all $\fs{T} \subset \fs{N}_t$
\[
    \Bigl|
        g(m';\fs{T}) - G(m';\fs{T})
    \Bigr|
    \leq C
\]
with probability $\geq 1-\delta$.

\begin{figure}[htbp]
\centering %
\begin{scriptsize}
\psfrag{or}{0} %
\psfrag{ca}{1} %
\psfrag{cb}{2} %
\psfrag{cc}{3} %
\psfrag{lx}{$t$} %
\psfrag{ly}{$d_t$} %
\includegraphics[width=7cm]{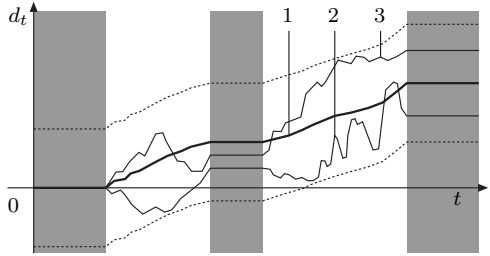}
\end{scriptsize}
\caption{If a divergence process is bounded, then the realizations (curves 2 \& 3) of a sub-divergence stay within a band around the mean (curve 1).}
\label{fig:divergence-boundedness} %
\end{figure}

Figure~\ref{fig:divergence-boundedness} illustrates this property. Boundedness is the key property that is going to be used to construct the results of this section. The first important result is that the posterior probability of the true input-output model is bounded from below.

\begin{theorem}\label{theo:lower-bound}
Let the set of operation modes of a controller be such that for all $m \in \fs{M}$ the divergence process $d_t(m^\ast\|m)$ is bounded. Then, for any $\delta > 0$, there is a $\lambda > 0$, such that for all $t \in \nats$,
\[
    P(m^\ast|\g{\hat{a}o}_{\leq t}) \geq \frac{\lambda}{|\fs{M}|}
\]
with probability $\geq 1 - \delta$.
\end{theorem}

\subsection{Core}

\begin{figure}[htbp]
\centering %
\begin{scriptsize}
\psfrag{h1}{$H_1$} \psfrag{p1}{$P_1$} %
\psfrag{h2}{$H_2$} \psfrag{p2}{$P_2$} %
\psfrag{h3}{$H_3$} \psfrag{p3}{$P_3$} %
\psfrag{ra1}{$A$} \psfrag{ra2}{$A$} %
\psfrag{rb1}{$B$} \psfrag{rb2}{$B$} %
\includegraphics[width=7cm]{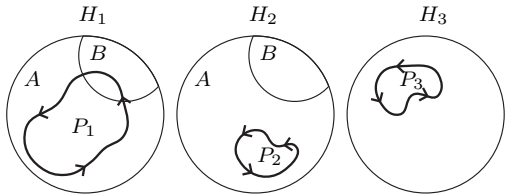}
\end{scriptsize}
\caption{If hypothesis $H_1$ is true and agrees with $H_2$ on region $A$, then
policy $P_2$ cannot disambiguate the three hypotheses.}
\label{fig:core} %
\end{figure}

If one wants to identify the operation modes whose posterior probabilities vanish, then it is not enough to characterize them as those modes whose hypothesis does not match the true hypothesis. Figure~\ref{fig:core} illustrates this problem. Here, three hypotheses along with their associated policies are shown. $H_1$ and $H_2$ share the prediction made for region~$A$ but differ in region~$B$. Hypothesis $H_3$ differs everywhere from the others. Assume $H_1$ is true. As long as we apply policy~$P_2$, hypothesis~$H_3$ will make wrong predictions and thus its divergence process will diverge as expected. However, no evidence against~$H_2$ will be accumulated. It is only when one applies policy~$P_1$ \emph{for long enough time} that the controller will eventually enter region~$B$ and hence accumulate counter-evidence for $H_2$.

But what does ``long enough'' mean? If $P_1$ is executed only for a short period, then the controller risks not visiting the disambiguating region. But unfortunately, neither the right policy nor the right length of the period to run it are known beforehand. Hence, an agent needs a clever time-allocating strategy to test all policies for all finite time intervals. This motivates the following definition.

The \emph{core} of an operation mode $m^\ast$, denoted as $[m^\ast]$, is the subset of $\fs{M}$ containing operation modes behaving like $m^\ast$ under its policy. More formally, an operation mode $m \notin [m^\ast]$ (i.e. is \emph{not} in the core) iff for any $C \geq 0$, $\delta, \xi > 0$, there is a $t_0 \in \nats$, such that for all $t \geq t_0$,
\[
    G(m^\ast;\fs{T}) \geq C
\]
with probability $\geq 1-\delta$, where $G(m^\ast;\fs{T})$ is a sub-divergence of $d_t(m^\ast\|m)$, and $\prob\{\tau \in \fs{T}\} \geq \xi$ for all $\tau \in \fs{N}_t$.

In other words, if the agent was to apply $m^\ast$'s policy in each time step with probability at least $\xi$, and under this strategy the expected sub-divergence $G(m^\ast;\fs{T})$ of $d_t(m^\ast\|m)$ grows unboundedly, then $m$ is not in the core of $m^\ast$. Note that demanding a strictly positive probability of execution in each time step guarantees that the agent will run $m^\ast$ for all possible finite time-intervals. As the following theorem shows, the posterior probabilities of the operation modes that are not in the core vanish almost surely.

\begin{theorem}\label{theo:vanishing-posterior}
Let the set of operation modes of an agent be such that for all $m \in \fs{M}$ the divergence process $d_t(m^\ast\|m)$ is bounded. If $m \notin [m^\ast]$, then $P(m|\g{\hat{a}o}_{\leq t}) \rightarrow 0$ as $t \rightarrow \infty$ almost surely.
\end{theorem}

\subsection{Consistency}

\begin{figure}[htbp]
\centering %
\begin{scriptsize}
\psfrag{h1}{$H_1$} \psfrag{p1}{$P_1$} %
\psfrag{h2}{$H_2$} \psfrag{p2}{$P_2$} %
\psfrag{ra1}{$A$} \psfrag{ra2}{$A$} %
\psfrag{rb1}{$B$} \psfrag{rb2}{$B$} %
\includegraphics[width=7cm]{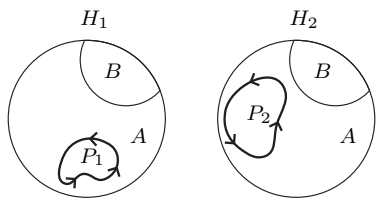}
\end{scriptsize}
\caption{An example of inconsistent policies. Both operation modes are in the core of each other, but have different policies.}
\label{fig:consistency} %
\end{figure}

Even if an operation mode $m$ is in the core of $m^\ast$, i.e. given that $m$ is essentially indistinguishable from $m^\ast$ under $m^\ast$'s control, it can still happen that $m^\ast$ and $m$ have different policies. Figure~\ref{fig:consistency} shows an example of this. The hypotheses~$H_1$ and~$H_2$ share region~$A$ but differ in region~$B$. In addition, both operation modes have their policies~$P_1$ and~$P_2$ respectively confined to region~$A$. Note that both operation modes are in the core of each other. However, their policies are different. This means that it is unclear whether multiplexing the policies in time will ever disambiguate the two hypotheses. This is undesirable, as it could impede the convergence to the right control law.

Thus, it is clear that one needs to impose further restrictions on the mapping of hypotheses into policies. With respect to Figure~\ref{fig:consistency}, one can make the following observations:

\begin{enumerate}
  \item Both operation modes have policies that select subsets of region~$A$.
  Therefore, the dynamics in~$A$ are preferred over the dynamics in~$B$.
  \item Knowing that the dynamics in~$A$ are preferred over the dynamics in~$B$
  allows us to drop region~$B$ from the analysis when choosing a policy.
  \item Since both hypotheses agree in region~$A$, \emph{they have to choose the same policy in order to be consistent in their selection criterion}.
\end{enumerate}

This motivates the following definition. An operation mode $m$ is said to be \emph{consistent} with $m^\ast$ iff $m \in [m^\ast]$ implies that for all $\varepsilon < 0$, there is a $t_0$, such that for all $t \geq t_0$ and all $\g{ao}_{<t}a_t$,
\[
    \Bigl|
    P(a_t|m^\ast,\g{ao}_{\leq t}) - P(a_t|m^\ast,\g{ao}_{\leq t})
    \Bigr|
    < \varepsilon.
\]

In other words, if $m$ is in the core of $m^\ast$, then $m$'s policy has to converge to $m^\ast$'s policy. The following theorem shows that consistency is a sufficient condition for convergence to the right control law.

\begin{theorem}\label{theo:convergence-bcr}
Let the set of operation modes of an agent be such that: for all $m \in \fs{M}$ the divergence process $d_t(m^\ast\|m)$ is bounded; and for all $m, m' \in \fs{M}$, $m$ is consistent with $m'$. Then,
\[
    P(a_t|\g{\hat{a}o}_{<t}) \rightarrow P(a_t|m^\ast,\g{ao}_{<t})
\]
almost surely as $t \rightarrow \infty$.
\end{theorem}

\subsection{Summary}

In this section, a proof of convergence of the Bayesian control rule to the true operation mode has been provided for a finite set of operation modes. For this convergence result to hold, two necessary conditions are assumed: boundedness and consistency.
The first one, \emph{boundedness}, imposes the stability of divergence processes under the partial influence of the policies contained within the set of operation modes. This condition can be regarded as an ergodicity assumption. The second one, \emph{consistency}, requires that if a hypothesis makes the same predictions as another hypothesis within its most relevant subset of dynamics, then both hypotheses share the same policy. This relevance is formalized as the \emph{core} of an operation mode.
The concepts and proof strategies strengthen the intuition about potential pitfalls that arise in the context of controller design. In particular we could show that the asymptotic analysis can be recast as the study of concurrent \emph{divergence processes} that determine the evolution of the posterior probabilities over operation modes, thus abstracting away from the details of the classes of I/O distributions.
The extension of these results to infinite sets of operation modes are left for future work. For example, one could think of partitioning a continuous space of operation modes into ``essentially different'' regions where representative operation modes subsume their neighborhoods \citep{Grunwald2007}.

\section{Examples}\label{sec:example}

\subsection{Bandit Problems}

Consider the \emph{multi-armed bandit problem} \citep{Robbins1952}.
The problem is stated as follows. Suppose there is an $N$-armed bandit, i.e. a slot-machine with $N$ levers. When pulled, lever $i$ provides a reward drawn from a Bernoulli distribution with a bias $h_i$ specific to that lever. That is, a reward $r=1$ is obtained with probability $h_i$ and a reward $r=0$ with probability $1-h_i$. The objective of the game is to maximize the time-averaged reward through iterative pulls. There is a continuum range of stationary strategies, each one parameterized by $N$ probabilities $\{s_i\}_{i=1}^N$ indicating the probabilities of pulling each lever. The difficulty arising in the bandit problem is to balance reward maximization based on the knowledge already acquired with attempting new actions to further improve knowledge. This dilemma is known as the exploration versus exploitation tradeoff \citep{Sutton1998}.

This is an ideal task for the Bayesian control rule, because each possible bandit has a known optimal agent. Indeed, a bandit can be represented by an $N$-dimensional bias vector $m=[m_1,\ldots,m_N] \in \fs{M} = [0;1]^N$. Given such a bandit, the optimal policy consists in pulling the lever with the highest bias. That is, an operation mode is given by:
\[
    P(o_t=1|m,a_t=i) = m_i
    \qquad
    P(a_t=i|m) =
    \begin{cases}
        1 & \text{if $i = \max_j \{m_j\}$,} \\
        0 & \text{else.}
    \end{cases}
\]
\begin{figure}[htbp]
    \centering
    \footnotesize
    \psfrag{aa}[c]{a)}
    \psfrag{ah1}[c]{$m_1$}
    \psfrag{ah2}[c]{$m_2$}
    \psfrag{ahh}[c]{$m_1 \geq m_2$}
    \psfrag{bb}[c]{b)}
    \psfrag{bh1}[c]{$m_1$}
    \psfrag{bh2}[c]{$m_2$}
    \psfrag{bh3}[c]{$m_3$}
    \psfrag{bhh}[c]{$m_2 \geq m_1,m_3$}
    \includegraphics[]{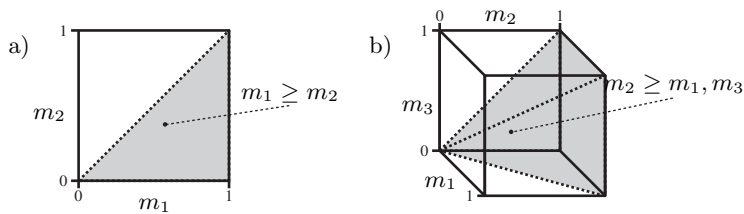}
    \caption{The space of bandit configurations can be partitioned
    into $N$ regions according to the optimal lever. Panel~a and~b
    show the 2-armed and 3-armed bandit cases respectively.}
    \label{fig:bandit-hypotheses}
\end{figure}

To apply the Bayesian control rule, it is necessary to fix a prior distribution over the bandit configurations. Assuming a uniform distribution, the Bayesian control rule is
\[
    P(a_{t+1}=i|\g{\hat ao}_{\leq t})
    = \int_{\fs{M}} P(a_{t+1}=i|m) P(m|\g{\hat ao}_{\leq t})
\]
with the update rule given by
\[
    P(m|\g{\hat ao}_{\leq t}) =
    \frac{
        P(m)\prod_{\tau=1}^t P(o_\tau|m,a_\tau)
    }{
        \int_{\fs{M}} P(m') \prod_{\tau=1}^t P(o_\tau|m',a_\tau) \,
        dm'
    }
    = \prod_{j=1}^N \frac{m_j^{r_j} (1-m_j)^{f_j}}{\betaf(r_j+1,f_j+1)}
\]
where $r_j$ and $f_j$ are the counts of the number of times a reward has been obtained from pulling lever $j$ and the number of times no reward was obtained respectively. Observe that here the summation over discrete operation modes has been replaced by an integral over the continuous space of configurations. In the last expression we see that the posterior distribution over the lever biases is given by a product of $N$ Beta distributions. Thus, sampling an action amounts to first sample an operation mode $m$ by obtaining each bias $m_j$ from a Beta distribution with parameters $r_j+1$ and $f_j+1$, and then choosing the action corresponding to the highest bias $i=\arg \max_j m_j$.

\begin{figure}[htbp]
\begin{center}
    \scriptsize
    \psfrag{ar}[c]{Avg. Reward}
    \psfrag{pe}[c]{\% Best Lever}
    \psfrag{m1}[l]{Bayesian control rule}
    \psfrag{m2}[l]{$\epsilon$-greedy}
    \psfrag{m3}[l]{Gittins indices}
    \includegraphics{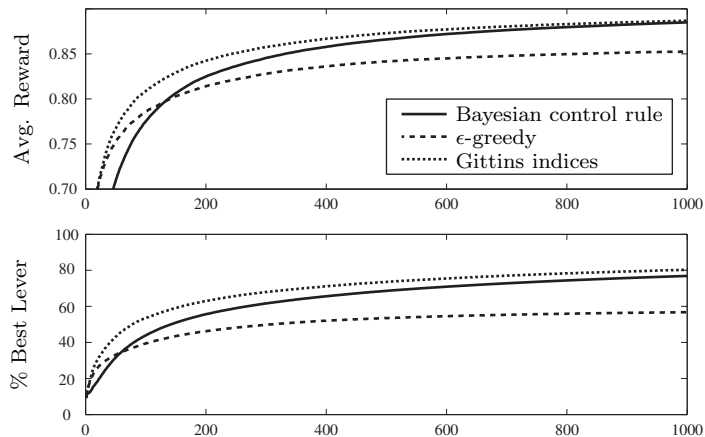}
    \caption{Comparison in the $N$-armed bandit problem
    of the Bayesian control rule (solid line),
    an $\varepsilon$-greedy agent (dashed line)
    and using Gittins indices (dotted line). 1,000 runs have
    been averaged. The top panel shows the evolution of the average
    reward. The bottom panel shows the evolution of the percentage
    of times the best lever was pulled.
    }\label{fig:bandit-comparison}
\end{center}\footnotesize
\end{figure}

\paragraph{Simulation:}
The Bayesian control rule described above has been compared against two other agents: an $\varepsilon$-greedy strategy with decay (on-line) and Gittins indices (off-line). The test bed consisted of bandits with $N=10$ levers whose biases were drawn uniformly at the beginning of each run. Every agent had to play 1000 runs for 1000 time steps each. Then, the performance curves of the individual runs were averaged. The $\varepsilon$-greedy strategy selects a random action with a small probability given by $\varepsilon \alpha^{-t}$ and otherwise plays the lever with highest expected reward. The parameters have been determined empirically to the values $\varepsilon=0.1$, $\alpha=0.99$ and $\tau=0.7$ after several test runs. They have been adjusted in a way to maximize the average performance in the last trials of our simulations. For the Gittins method, all the indices were computed up to horizon 1300 using a geometric discounting of $\alpha=0.999$, i.e. close to one to approximate the time-averaged reward. The results are shown in Figure~\ref{fig:bandit-comparison}.

It is seen that $\varepsilon$-greedy strategy quickly reaches an acceptable level of performance, but then seems to stall at a significantly suboptimal level, pulling the optimal lever only 60\% of the time. This can be improved by using a value for $\varepsilon$ that decays over time. In contrast, both the Gittins strategy and the Bayesian control rule show essentially the same asymptotic performance, but differ in the initial transient phase where the Gittins strategy significantly outperforms the Bayesian control rule. There are at least three observations that are worth making here. First, Gittins indices have to be pre-computed off-line. The time complexity scales quadratically with the horizon, and the computations for the horizon of 1300 steps took several hours on our machines. In contrast, the Bayesian control rule could be applied without pre-computation. Second, even though the Gittins method actively issues the optimal information gathering actions while the Bayesian control rule passively samples the actions from the posterior distribution over operation modes, in the end both methods rely on the convergence of the underlying Bayesian estimator. This implies that both methods have the same information bottleneck, since the Bayesian estimator requires the same amount of information to converge. Thus, active information gathering actions only affect the utility of the transient phase, not the permanent state. Other efficient algorithms for bandit problems can be found
in the literature \citep{Auer2002}.

\subsection{Markov Decision Problems}

\label{subsec:MDP}

A Markov Decision Process (\emph{MDP}) is defined as a tuple $(\fs{X},\fs{A},T,r)$: $\fs{X}$ is the
state space; $\fs{A}$ is the action space; $T_a(x;x')=\prob(x'|a,x)$ is the
probability that an action $a \in \fs{A}$ taken in state $x \in \fs{X}$ will
lead to state $x'\in \fs{X}$; and $r(x,a) \in \fs{R} \define \reals$ is the
immediate reward obtained in state $x \in \fs{X}$ and action $a \in \fs{A}$.
The interaction proceeds in time steps $t=1,2,\ldots$ where at time $t$, action
$a_t \in \fs{A}$ is issued in state $x_{t-1} \in \fs{X}$, leading to a reward
$r_t = r(x_{t-1}, a_t)$ and a new state $x_t$ that starts the next time step
$t+1$. A stationary closed-loop control policy $\pi : \fs{X} \rightarrow \fs{A}$
assigns an action to each state. For MDPs there always exists an optimal
stationary deterministic policy and thus one only needs to consider such
policies. In undiscounted MDPs the average reward per time step for a fixed policy $\pi$
with initial state $x$ is defined as $\rho^\pi(x) = \lim_{t\rightarrow
\infty} \expect^\pi[\frac{1}{t}\sum_{\tau=0}^t r_\tau]$. It can be shown
\citep{Bertsekas1987} that $\rho^\pi(x) = \rho^\pi(x')$ for all $x, x' \in
\fs{X}$ under the assumption that the Markov chain for policy $\pi$ is ergodic.
Here, we assume that the MDPs are ergodic for all stationary policies.
Following the Q-notation of \citet{Watkins1989}, the optimal policy $\pi^\ast$
can be characterized in terms of the optimal average reward $\rho$ and the
optimal relative Q-values $Q(x,a)$ for each state-action pair $(x,a)$ that are
solutions to the following system of non-linear equations \citep{Singh1994}:
for any state $x \in \fs{X}$ and action $a \in \fs{A}$,
\begin{equation}\label{eq:bellman-equations}
\begin{aligned}
    Q(x,a) + \rho
        &= r(x,a) + \sum_{y \in \fs{X}} \prob(x'|x,a) \Bigl[
        \max_{a'} Q(x',a') \Bigr]
        \\
        &= r(x,a) + \expect_{x'}\Bigl[ \max_{a'} Q(x',a') \Bigr| x, a\Bigr].
\end{aligned}
\end{equation}
The optimal policy can then be defined as $\pi^\ast(x) \define \arg
\max_{a}Q(x,a)$ for any state $x \in \fs{X}$.

Again this setup allows
for a straightforward solution with the Bayesian control rule, because
each possible MDP (characterized by the Q-values and the average reward)
has a known solution $\pi^\ast(x)$. Accordingly, the operation modes $m$ are
given by $(Q_m,\rho_m)$. To obtain a likelihood model for inference over $m$,
we realize that equation~(\ref{eq:bellman-equations}) can be rewritten such
that it predicts the instantaneous reward $r(x,a)$ as the sum of a
mean instantaneous reward $\xi_m$ plus a noise term $\nu$ given the $Q_m$-values and the average
reward~$\rho_m$ for the MDP labeled by $m$
\[
r(x,a) = \underbrace{Q_m(x,a) + \rho_m - \max_{a'} Q_m(x',a')}_{\text{mean instantaneous reward } \xi_m(x,a,x')} + \underbrace{\max_{a'} Q_m(x',a') - \expect[\max_{a'} Q_m(x',a')|x,a]}_{\text{noise } \nu}
\]
Assuming that $\nu$ can be reasonably approximated by
a normal distribution $\normal(0, 1/p)$ with precision $p$, we can write
down a likelihood model for the immediate reward $r$ using the Q-values and the
average reward, i.e.
\begin{equation}\label{eq:likelihood-reward}
    P(r|m,x,a,x') = \sqrt{\frac{p}{2\pi}} \exp\Bigl\{
        - \frac{p}{2} (r-\xi_m(x,a,x'))^2
        \Bigr\}.
\end{equation}
In order to determine the intervention model for each operation mode,
we can simply exploit the above properties of the $Q$-values, which gives
\begin{align}
    \label{eq:prob-action}
    P(a|m, x)
    = \begin{cases}
        1 & \text{if $a=\arg\max_{a'} Q(x,a')$} \\
        0 & \text{else.}
    \end{cases}
\end{align}
To apply the Bayesian control rule over the controllers $m$, the
intervened posterior distribution $P(m|\hat{a}_{\leq t}, x_{\leq t})$
needs to be computed. Fortunately,
due to the simplicity of the likelihood model, one can easily devise a
conjugate prior distribution and apply standard inference methods (see Appendix~\ref{subsec:gibbs}).
Actions are again determined by sampling operation modes from this posterior
and executing the action suggested by the corresponding intervention models.
The resulting algorithm is very similar to Bayesian Q-learning \citep{Dearden1998,Dearden1999},
but differs in the way actions are selected.

\begin{figure*}[htbp]
\begin{center}
    \small\psfrag{aa}[c]{a)}
    \includegraphics[scale=0.93]{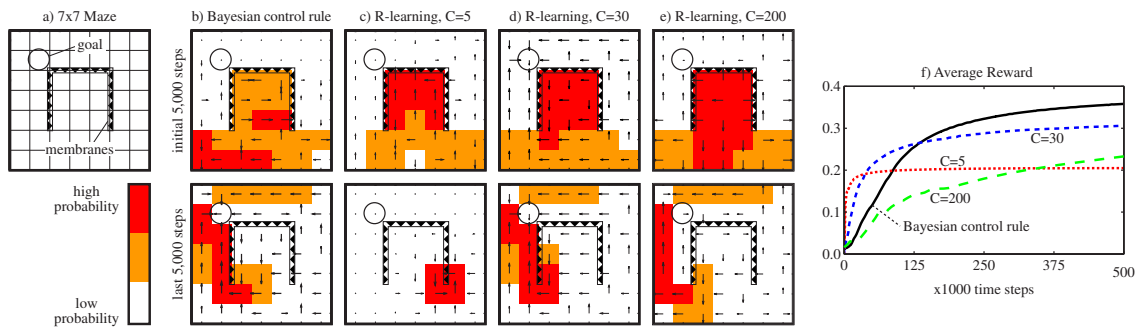}
    \caption{Results for the $7 \times 7$ grid-world domain. Panel (a)
    illustrates the setup. Columns (b)-(e) illustrate the behavioral statistics
    of the algorithms. The upper and lower row have been calculated
    over the first and last 5,000 time steps of randomly chosen
    runs. The probability of being in a state is color-encoded, and
    the arrows represent the most frequent actions taken by the agents.
    Panel (f) presents the curves obtained by averaging ten runs.
    }
    \label{fig:7x7maze}
\end{center}
\end{figure*}

\paragraph{Simulation:}
We have tested our MDP-agent in a grid-world example.
To give an intuition of the achieved performance,
the results are contrasted with those achieved by R-learning. We have used the
R-learning variant presented in \citet[Algorithm 3]{Singh1994} together with
the uncertainty exploration strategy \citep{Mahadevan1996}. The corresponding
update equations are
\begin{equation}
\begin{aligned}
    Q(x,a) &\leftarrow (1-\alpha)Q(x,a) + \alpha \bigl( r-\rho+\max_{a'}Q(x',a') \bigr) \\
    \rho &\leftarrow (1-\beta)\rho + \beta \bigl( r + \max_{a'} Q(x',a')
    -Q(x,a) \bigr),
\end{aligned}
\end{equation}
where $\alpha, \beta > 0$ are learning rates. The exploration strategy chooses
with fixed probability $p_{\exp}>0$ the action $a$ that maximizes $Q(x,a) +
\frac{C}{F(x,a)}$, where $C$ is a constant, and $F(x,a)$ represents the number
of times that action $a$ has been tried in state $x$. Thus, higher values of
$C$ enforce increased exploration.

In \citet{Mahadevan1996}, a grid-world is described that is especially useful
as a test bed for the analysis of RL algorithms. For our purposes, it is of
particular interest because it is easy to design experiments containing
\emph{suboptimal limit-cycles}.
Figure~\ref{fig:7x7maze}, panel (a), illustrates the $7\times7$ grid-world. A
controller has to learn a policy that leads it from any initial location to the
goal state. At each step, the agent can move to any adjacent space (up, down,
left or right). If the agent reaches the goal state then its next position is
randomly set to any square of the grid (with uniform probability) to start
another trial. There are also ``one-way membranes'' that allow the agent to
move into one direction but not into the other. In these experiments, these
membranes form ``inverted cups'' that the agent can enter from any side but can
only leave through the bottom, playing the role of a local maximum. Transitions
are stochastic: the agent moves to the correct square with probability
$p=\frac{9}{10}$ and to any of the free adjacent spaces (uniform distribution)
with probability $1-p=\frac{1}{10}$. Rewards are assigned as follows. The
default reward is $r=0$. If the agent traverses a membrane it obtains a reward
of $r=1$. Reaching the goal state assigns $r=2.5$.
The parameters chosen for this simulation were the following. For our MDP-agent, we
have chosen hyperparameters $\mu_0 = 1$ and $\lambda_0 = 1$ and precision
$p=1$. For R-learning, we have chosen learning rates $\alpha=0.5$ and
$\beta=0.001$, and the exploration constant has been set to $C=5$, $C=30$ and to
$C=200$. A total of 10 runs were carried out for each algorithm. The results are
presented in Figure~\ref{fig:7x7maze} and Table~\ref{tab:7x7maze}. R-learning
only learns the optimal policy given sufficient exploration (panels c \& d,
bottom row), whereas the Bayesian control rule learns the policy successfully. In
Figure~\ref{fig:7x7maze}e, the learning curve of R-learning for $C=5$ and $C=30$ is initially
steeper than the Bayesian controller. However, the latter attains a higher
average reward around time step 125,000 onwards. We attribute this shallow
initial transient to the phase where the distribution over the operation modes
is flat, which is also reflected by the initially random exploratory behavior.

\begin{table}
\label{tab:7x7maze}
\begin{center}
\begin{footnotesize}
\begin{tabular}{lc}
    \toprule
    & Average Reward \\
    \midrule
    BCR               & $0.3582 \pm 0.0038$ \\
    R-learning, $C=200$   & $0.2314 \pm 0.0024$ \\
    R-learning, $C=30$    & $0.3056 \pm 0.0063$ \\
    R-learning, $C=5$     & $0.2049 \pm 0.0012$ \\
    \bottomrule
\end{tabular}
\end{footnotesize}
\end{center}
\caption{Average reward attained by the different algorithms at the end of the
run. The mean and the standard deviation has been calculated based on 10
runs.}
\end{table}

\section{Discussion}\label{sec:discussion}

The key idea of this work is to extend the minimum relative entropy principle, i.e. the variational principle underlying Bayesian estimation, to the problem of adaptive control. From a coding point of view, this work extends the idea of maximal compression of the observation stream to the whole experience of the agent containing both the agent's actions and observations. This not only minimizes the amount of bits to write when \emph{saving/encoding} the I/O stream, but it also minimizes the amount of bits required to \emph{produce/decode} an action \citep[Chapter 6]{Mackay2003}.

This extension is non-trivial, because there is an important caveat for coding I/O sequences: unlike observations, actions do not carry any information that could be used for inference in adaptive coding because actions are issued by the decoder itself. The problem is that doing inference on ones own actions is logically inconsistent and leads to paradoxes \citep{Nozick1969}. This seemingly innocuous issue has turned out to be very intricate and has been investigated intensely in the recent past by researchers focusing on the issue of causality \citep{Pearl2000, Spirtes2000, Dawid2010}. Our work contributes to this body of research by providing further evidence that actions cannot be treated using probability calculus alone.

If the causal dependencies are carefully taken into account, then minimizing the relative entropy leads to a rule for adaptive control which has been called the Bayesian control rule. This rule allows combining a class of task-specific agents into an agent that is universal with respect to this class. The resulting control law is a simple stochastic control rule that is completely general and parameter-free. As the analysis in this paper shows, this control rule converges to the true control law under mild assumptions.

\subsection{Critical issues}

\begin{itemize}
    \item \emph{Causality.}
    Virtually every adaptive control method in the literature successfully treats actions as conditionals over observation streams and never worries about causality. Thus, why bother about interventions? In a decision-theoretic setup, the decision maker chooses a policy $\pi^\ast \in \Pi$ maximizing the expected utility $U$ over the outcomes $\omega \in \Omega$, i.e. $\pi^\ast \define \arg \max_\pi \expect[U|\pi] = \sum_\omega \prob(\omega|\pi) U(\omega)$. ``Choosing $\pi^\ast$'' is formally equivalent to choosing the Kronecker delta function $\delta_{\pi^\ast}^\pi$ as the probability distribution over policies. In this case, the conditional probabilities $\prob(\omega|\pi)$ and $\prob(\omega|\hat{\pi})$ coincide, since
    \[
        \prob(\omega, \pi)
        = \prob(\pi) \prob(\omega|\pi)
        = \delta_{\pi^\ast}^\pi \prob(\omega|\pi)
        = \prob(\omega, \hat{\pi}).
    \]
    Hence, the formalization of actions as interventions and observations as conditions is perfectly compatible with the decision-theoretic setup and in fact generalizes decision variables to the status of intervened random variables.
    \item \emph{Where do prior probabilities/likelihood models/policies come from?}
    The predictor in the Bayesian control rule is essentially a Bayesian predictor and thereby entails (almost) the same modeling paradigm. The designer has to define a class of hypotheses over the environments, construct appropriate likelihood models, and choose a suitable prior probability distribution to capture the model's uncertainty. Similarly, under sufficient domain knowledge, an analogous procedure can be applied to construct suitable operation modes. However, there are many situations where this is a difficult or even intractable problem in itself. For example, one can design a class of operation modes by pre-computing the optimal policies for a given class of environments. Formally, let $\Theta$ be a class of hypotheses modeling environments and let $\Pi$ be class of policies. Given a utility criterion $U$, define the set of operation modes $\fs{M} \define \{m_\theta\}_{\theta \in \Theta}$ by constructing each operation mode as $m_\theta \define (\theta,\pi^\ast)$, $\pi^\ast \in \pi$, where $\pi^\ast \define \arg\max_\pi \expect[U|\theta,\pi]$. However, computing the optimal policy $\pi^\ast$ is in many cases intractable. In some cases, this can be remedied by characterizing the operation modes through optimality equations which are solved by probabilistic inference as in the example of the MDP agent in Section~\ref{subsec:MDP}. Recently, we have applied a similar approach to adaptive control problems with linear quadratic regulators \citep{OrtegaBraun2010c}.
    \item \emph{Problems of Bayesian methods.}
    The Bayesian control rule treats an adaptive control problem as a Bayesian inference problem. Hence, all the problems typically associated with Bayesian methods carry over to agents constructed with the Bayesian control rule. These problems are of both analytical and computational nature. For example, there are many probabilistic models where the posterior distribution does not have a closed-form solution. Also, exact probabilistic inference is in general computationally very intensive. Even though there is a large literature in efficient/approximate inference algorithms for particular problem classes \citep{Bishop2006}, not many of them are suitable for on-line probabilistic inference in more realistic environment classes.
    \item \emph{Bayesian control rule versus Bayes-optimal control.} Directly maximizing the (subjective) expected utility for a given environment class is not the same as minimizing the expected relative entropy for a given class of operation modes. As such, the Bayesian control rule is not a Bayes-optimal controller. Indeed, it is easy to design experiments where the Bayesian control rule converges exponentially slower (or does not converge at all) than a Bayes-optimal controller to the maximum utility. Consider the following simple example: Environment 1 is a $k$-state MDP in which only $k$ consecutive actions $A$ reach a state with reward $+1$. Any interception with a $B$-action leads back to the initial state. Consider a second environment which is like the first but actions $A$ and $B$ are interchanged. A Bayes-optimal controller figures out the true environment in $k$ actions (either $k$ consecutive $A$'s or $B$'s). Consider now the Bayesian control rule: The optimal action in Environment~1 is $A$, in Environment~2 is $B$. A uniform ($\frac{1}{2},\frac{1}{2}$) prior over the operation modes stays a uniform posterior as long as no reward has been observed. Hence the Bayesian control rule chooses at each time-step $A$ and $B$ with equal probability. With this policy it takes about $2^k$ actions to accidentally choose a row of $A$'s (or $B$'s) of length $k$. From then on the Bayesian control rule is optimal too. So a Bayes-optimal controller converges in time $k$, while the Bayesian control rule needs exponentially longer. One way to remedy this problem might be to allow the Bayesian control rule to sample actions from the same operation mode for several time steps in a row rather than randomizing controllers in every cycle. However, if one considers non-stationary environments this strategy can also break down. Consider, for example, an increasing MDP with $k = \bigl\lceil 10\sqrt{t} \,\bigr\rceil$, in which a Bayes-optimal controller converges in $100$ steps, while the Bayesian control rule does not converge at all in most realizations, because the boundedness
        assumption is violated.
\end{itemize}

\subsection{Relation to existing approaches}

Some of the ideas underlying this work are not unique to the Bayesian control rule. The following is a selection of previously published work in the recent Bayesian reinforcement learning literature where related ideas can be found.

\begin{itemize}
    \item \emph{Compression principles.}
    In the literature, there is an important amount of work relating compression to intelligence \citep{Mackay2003, Hutter2004}. In particular, it has been even proposed that compression ratio is an objective quantitative measure of intelligence \citep{Mahoney1999}. Compression has also been used as a basis for a theory of curiosity, creativity and beauty \citep{Schmidhuber2009}.
    \item \emph{Mixture of experts.}
    Passive sequence prediction by mixing experts has been studied extensively in the literature \citep{CesaBianchi2006}. In \citep{Hutter2004b}, Bayes-optimal predictors are mixed. Bayes-mixtures
    can also be used for universal prediction \citep{Hutter2003}.
    For the control case, the idea of using mixtures of expert-controllers has been previously evoked in models like the MOSAIC-architecture \citep{Haruno2001}. Universal learning with Bayes mixtures of experts in reactive environments has been studied in \citep{PolandHutter2005,Hutter2002}.
    \item \emph{Stochastic action selection.}
    Other stochastic action selection approaches are found in \citet{Wyatt1997} who examines exploration strategies for (PO)MDPs, in learning automata \citep{Narendra1974} and in probability matching \citep{DudaHartStork2000} amongst others. In particular, \citet{Wyatt1997} discusses theoretical properties of an extension to \emph{probability matching} in the context of multi-armed bandit problems. There, it is proposed to choose a lever according to how likely it is to be optimal and it is shown that this strategy converges, thus providing a simple method for guiding exploration.
    \item \emph{Relative entropy criterion.}
    The usage of a minimum relative entropy criterion to derive control laws underlies the KL-control methods developed in \citet{Todorov2006, Todorov2009, Kappen2009}. There, it has been shown that a large class of optimal control problems can be solved very efficiently if the problem statement is reformulated as the minimization of the deviation of the dynamics of a controlled system from the uncontrolled system. A related idea is to conceptualize planning as an inference problem \citep{Toussaint2006}. This approach is based on an equivalence between maximization of the expected future return and likelihood maximization which is both applicable to MDPs and POMDPs. Algorithms based on this duality have become an active field of current research. See for example \citet{Rasmussen2008}, where very fast model-based RL techniques are used for control in continuous state and action spaces.
\end{itemize}

\section{Conclusions}\label{sec:conclusions}

This work introduces the Bayesian control rule, a Bayesian rule for adaptive control. The key feature of this rule is the special treatment of actions based on causal calculus and the decomposition of an adaptive agent into a mixture of operation modes, i.e. environment-specific agents. The rule is derived by minimizing the expected relative entropy from the true operation mode and by carefully distinguishing between actions and observations. Furthermore, the Bayesian control rule turns out to be exactly the predictive distribution over the next action given the past interactions that one would obtain by using only probability and causal calculus. Furthermore, it is shown that agents constructed with the Bayesian control rule converge to the true operation mode under mild assumptions: boundedness, which related to ergodicity; and consistency, demanding that two indistinguishable hypotheses share the same policy.

We have presented the Bayesian control rule as a way to solve adaptive control problems based on a
minimum relative entropy principle. Thus, the Bayesian control rule can either be regarded as a new principled approach to adaptive control under a novel optimality criterion or as a heuristic approximation to traditional Bayes-optimal control. Since it takes on a similar form to Bayes' rule, the adaptive control problem could then be translated into an on-line inference problem where actions are sampled stochastically from a posterior distribution. It is important to note, however, that the problem statement as formulated here and the usual Bayes-optimal approach in adaptive control are \emph{not} the same. In the future the relationship between these two problem statements deserves further investigation.

\newpage

% Acknowledgements
\acks{We thank David Wingate, Zoubin Ghahramani, Jos\'e Aliste, Jos\'e Donoso, Humberto Maturana and the anonymous reviewers for comments on earlier versions of this manuscript and/or inspiring discussions. We thank the Ministerio de Planificaci\'{o}n de Chile (MIDEPLAN) and the B\"{o}hringer-Ingelheim-Fonds (BIF) for funding.}

%%%%%%%%%%%%%%%%%%%%%%%%%%%%%%%%%%%%%%%%%%%%%%%%%%%%%%%%%%%%%%%%%%%%
% Appendix                                                         %
%%%%%%%%%%%%%%%%%%%%%%%%%%%%%%%%%%%%%%%%%%%%%%%%%%%%%%%%%%%%%%%%%%%%
\appendix

\section{Proofs}

\subsection{Proof of Theorem~\ref{theo:minimum-total-divergence}}
\begin{proof}
The proof follows the same line of argument as the solution to Equation~\ref{eq:minimization-bayesian} with the crucial difference that actions are treated as interventions. Consider without loss of generality the summand $\sum_m P(m) C^{a_t}_m$ in Equation~\ref{eq:total-causal-divergence}. Note that the relative entropy can be written as a difference of two logarithms, where only one term depends on $\prob$ to be varied. Therefore, one can integrate out the other term and write it as a constant $c$. This yields
\[
    c - \sum_{m} \, P(m)
    \sum_{\g{ao}_{<t}} \, P(\g{\hat{a}o}_{<t}|m) \,
        \sum_{a_t} P(a_t|m,\g{\hat{a}o}_{<t})
        \ln \prob(a_t|\g{ao}_{<t}).
\]
Substituting $P(\g{\hat ao}_{<t}|m)$ by $P(m|\g{\hat ao}_{<t}) P(\g{\hat ao}_{<t}) / P(m)$ using Bayes' rule and further rearrangement of the terms leads to
\begin{align*}
    &= c - \sum_{m} \sum_{\g{ao}_{<t}}
        \, P(m|\g{\hat{a}o}_{<t}) P(\g{\hat{a}o_{<t}}) \,
        \sum_{a_t} P(a_t|m,\g{\hat{a}o}_{<t})
        \ln \prob(a_t|\g{ao}_{<t}) \\
    &= c - \sum_{\g{ao}_{<t}} \, P(\g{\hat{a}o}_{<t}) \,
        \sum_{a_t} P(a_t|\g{\hat{a}o}_{<t})
        \ln \prob(a_t|\g{ao}_{<t}).
\end{align*}
The inner sum has the form $-\sum_x p(x) \ln q(x)$, i.e. the cross-entropy between $q(x)$ and $p(x)$, which is minimized when $q(x)=p(x)$ for all $x$. By choosing this optimum one obtains $\prob(a_t|\g{ao}_{<t}) = P(a_t|\g{\hat{a}o}_{<t})$ for all $a_t$. Note that the solution to this variational problem is independent of the weighting $P(\g{\hat{a}o}_{<t})$. Since the same argument applies to any summand $\sum_m P(m) C^{a_\tau}_m$ and $\sum_m P(m) C^{o_\tau}_m$ in Equation~\ref{eq:total-causal-divergence}, their variational problems are mutually independent. Hence,
\[
    \agent(a_t|\g{ao}_{<t}) = P(a_t|\g{\hat{a}o}_{<t})
    \qquad
    \agent(o_t|\g{ao}_{<t}) = P(o_t|\g{\hat{a}o}_{<t}\hat{a}_t)
\]
for all $\g{ao}_{\leq t} \in \finint$. For $P(a_t|\g{\hat{a}o}_{<t})$, introduce the variable $m$ via a marginalization and then apply the chain rule:
\[
    P(a_t|\g{\hat{a}o}_{<t})
    = \sum_{m} P(a_{t+1}|m,\g{\hat{a}o}_{<t})
        P(m|\g{\hat{a}o}_{<t}).
\]
The term $P(m|\g{\hat ao}_{\leq t})$ can be further developed as
\begin{align*}
    P(m|\g{\hat{a}o}_{<t})
    &= \frac{P(\g{\hat{a}o}_{<t}|m) P(m)}
        {\sum_{m'} P(\g{\hat{a}o}_{<t}|m') P(m')}
    \\
    &= \frac{ P(m) \prod_{\tau=1}^{t-1}
        P(\hat{a}_\tau|m,\g{\hat{a}o}_{<\tau})
        P(o_\tau|m,\g{\hat{a}o}_{<\tau}\hat{a}_\tau) }
        { \sum_{m'} P(m') \prod_{\tau=1}^{t-1}
        P(\hat{a}_\tau|m',\g{\hat{a}o}_{<\tau})
        P(o_\tau|m',\g{\hat{a}o}_{<\tau}\hat{a}_\tau) }
    \\
    &= \frac{ P(m) \prod_{\tau=1}^{t-1}
        P(o_\tau|m,\g{ao}_{<\tau}a_\tau) }
        { \sum_{m'} P(m') \prod_{\tau=1}^{t-1}
        P(o_\tau|m',\g{ao}_{<\tau}a_\tau) }.
\end{align*}
The first equality is obtained by applying Bayes' rule and the
second by using the chain rule for probabilities. The second equality follows from using the causal factorization of the joint probability distribution. To get the last equality, one applies the interventions to the causal factorization. Thus, $P(\hat{a}_\tau|m,\g{\hat{a}o}_{<\tau}) = 1$ and $P(o_\tau|m,\g{\hat{a}o}_{<\tau}\hat{a}_\tau) = P(o_\tau|m,\g{ao}_{<\tau}a_\tau)$. The equations characterizing $P(o_t|\g{\hat{a}o}_{<t}\hat{a}_t)$ are obtained similarly.
\end{proof}

\subsection{Proof of Theorem~\ref{theo:lower-bound}}
\begin{proof}
As has been pointed out in~(\ref{eq:divergence-decomposition}), a particular realization of the divergence process $d_t(m^\ast\|m)$ can be decomposed as
\[
    d_t(m^\ast\|m)
    = \sum_{m'} g_m(m';\fs{T}_{m'}),
\]
where the $g_m(m';\fs{T}_{m'})$ are sub-divergences of $d_t(m^\ast\|m)$ and the $\fs{T}_{m'}$ form a partition of $\fs{N}_t$. However, since $d_t(m^\ast\|m)$ is bounded in $\fs{M}$, one has for all $\delta'>0$, there is a $C(m) \geq 0$, such that for all $m' \in \fs{M}$, all $t \in \fs{N}_t$ and all $\fs{T} \subset \fs{N}_t$, the inequality
\[
    \Bigl|
        g_m(m';\fs{T}_{m'}) - G_m(m';\fs{T}_{m'})
    \Bigr|
    \leq C(m)
\]
holds with probability $\geq 1-\delta'$. However, due to~(\ref{eq:gibbs-ineq-sub-diver}),
\[
    G_m(m';\fs{T}_{m'}) \geq 0
\]
for all $m' \in \fs{M}$. Thus,
\[
    g_m(m';\fs{T}_{m'}) \geq -C(m).
\]
If all the previous inequalities hold simultaneously then the divergence process can be bounded as well. That is, the inequality
\begin{equation}\label{eq:proof-1-a}
    d_t(m^\ast\|m) \geq -M C(m)
\end{equation}
holds with probability $\geq (1-\delta')^M$ where $M \define |\fs{M}|$. Choose
\[
    \beta(m) \define \max\{0, \ln \tfrac{P(m)}{P(m^\ast)}\}.
\]
Since $0 \geq \ln \tfrac{P(m)}{P(m^\ast)} - \beta(m)$, it can be added to the right hand side of~(\ref{eq:proof-1-a}). Using the definition of $d_t(m^\ast\|m)$, taking the exponential and rearranging the terms one obtains
\[
    P(m^\ast) \prod_{\tau=1}^t P(o_\tau|m^\ast,\g{ao}_{<\tau}a_\tau)
    \geq e^{-\alpha(m)} P(m) \prod_{\tau=1}^t P(o_\tau|m^\ast,\g{ao}_{<\tau}a_\tau)
\]
where $\alpha(m) \define M C(m) + \beta(m) \geq 0$. Identifying the posterior probabilities of $m^\ast$ and $m$ by dividing both sides by the normalizing constant yields the inequality
\[
    P(m^\ast|\g{\hat{a}o}_{\leq t}) \geq e^{-\alpha(m)} P(m|\g{\hat{a}o}_{\leq t}).
\]
This inequality holds simultaneously for all $m \in \fs{M}$ with probability $\geq (1-\delta')^{M^2}$ and in particular for $\lambda \define \min_m \{e^{-\alpha(m)}\}$, that is,
\[
    P(m^\ast|\g{\hat{a}o}_{\leq t}) \geq \lambda P(m|\g{\hat{a}o}_{\leq t}).
\]
But since this is valid for any $m \in \fs{M}$, and because $\max_m \{ P(m|\g{\hat{a}o}_{\leq t}) \} \geq \frac{1}{M}$, one gets
\[
    P(m^\ast|\g{\hat{a}o}_{\leq t}) \geq \frac{\lambda}{M},
\]
with probability $\geq 1-\delta$ for arbitrary $\delta > 0$ related to $\delta'$ through the equation $\delta' \define 1 - \sqrt[M^2]{1-\delta}$.
\end{proof}

\subsection{Proof of Theorem~\ref{theo:vanishing-posterior}}

\begin{proof}
The divergence process $d_t(m^\ast\|m)$ can be decomposed into a sum of sub-divergences (see Equation~\ref{eq:divergence-decomposition})
\begin{equation}\label{eq:proof-2-a}
    d_t(m^\ast\|m) = \sum_{m'} g(m';\fs{T}_{m'}).
\end{equation}
Furthermore, for every $m' \in \fs{M}$, one has that for all $\delta > 0$, there is a $C \geq 0$, such that for all $t \in \nats$ and for all $\fs{T} \subset \fs{N}_t$
\[
    \Bigl|
        g(m';\fs{T}) - G(m';\fs{T})
    \Bigr|
    \leq C(m)
\]
with probability $\geq 1-\delta'$. Applying this bound to the summands in~(\ref{eq:proof-2-a}) yields the lower bound
\[
    \sum_{m'} g(m';\fs{T}_{m'})
    \geq \sum_{m'} \bigl( G(m';\fs{T}_{m'}) - C(m) \bigr)
\]
which holds with probability $\geq (1-\delta')^M$, where $M \define |\fs{M}|$. Due to Inequality~\ref{eq:gibbs-ineq-sub-diver}, one has that for all $m' \neq m^\ast$, $G(m';\fs{T}_{m'}) \geq 0$. Hence,
\[
    \sum_{m'} \bigl( G(m';\fs{T}_{m'}) - C(m) \bigr)
    \geq G(m^\ast;\fs{T}_{m^\ast}) - M C
\]
where $C \define \max_{m} \{C(m)\}$. The members of the set $\fs{T}_{m^\ast}$ are determined stochastically; more specifically, the $i^\text{th}$ member is included into $\fs{T}_{m^\ast}$ with probability $P(m^\ast|\g{\hat{a}o}_{\leq i})$. But since $m \notin [m^\ast]$, one has that $G(m^\ast;\fs{T}_{m^\ast}) \rightarrow \infty$ as $t \rightarrow \infty$ with probability $\geq 1-\delta'$ for arbitrarily chosen $\delta'>0$. This implies that
\[
    \lim_{t \rightarrow \infty} d_t(m^\ast\|m)
    \geq \lim_{t \rightarrow \infty} G(m^\ast;\fs{T}_{m^\ast}) - MC
    \nearrow \infty
\]
with probability $\geq 1 - \delta$, where $\delta>0$ is arbitrary and related to $\delta'$ as $\delta = 1-(1-\delta')^{M+1}$. Using this result in the upper bound for posterior probabilities yields the final result
\[
    0 \leq \lim_{t \rightarrow \infty} P(m|\g{\hat{a}o}_{\leq t})
    \leq \lim_{t \rightarrow \infty} \frac{P(m)}{P(m^\ast)}
        e^{-d_t(m^\ast\|m)}
    = 0.
\]
\end{proof}

\subsection{Proof of Theorem~\ref{theo:convergence-bcr}}

\begin{proof}
We will use the abbreviations $p_{m}(t) \define P(a_t|m, \g{\hat{a}o}_{<t})$ and $w_m(t) \define P(m|\g{\hat{a}o}_{<t})$. Decompose $P(a_t|\g{\hat{a}o}_{<t})$ as
\begin{equation}\label{eq:proof-3-a}
    P(a_t|\g{\hat{a}o}_{<t})
    = \sum_{m \notin [m^\ast]} p_m(t) w_m(t)
    + \sum_{m \in [m^\ast]} p_m(t) w_m(t).
\end{equation}

The first sum on the right-hand side is lower-bounded by zero and upper-bounded by
\[
    \sum_{m \notin [m^\ast]} p_m(t) w_m(t)
    \leq \sum_{m \notin [m^\ast]} w_m(t)
\]
because $p_m(t) \leq 1$. Due to Theorem~\ref{theo:vanishing-posterior}, $w_m(t) \rightarrow 0$ as $t \rightarrow \infty$ almost surely. Given $\varepsilon'>0$ and $\delta'>0$, let $t_0(m)$ be the time such that for all $t \geq t_0(m)$, $w_m(t) < \varepsilon'$. Choosing $t_0 \define \max_m \{t_0(m)\}$, the previous inequality holds for all $m$ and $t \geq t_0$ simultaneously with probability $\geq (1-\delta')^M$. Hence,
\begin{equation}\label{eq:proof-3-b}
    \sum_{m \notin [m^\ast]} p_m(t) w_m(t)
    \leq \sum_{m \notin [m^\ast]} w_m(t)
    < M \varepsilon'.
\end{equation}

To bound the second sum in~(\ref{eq:proof-3-a}) one proceeds as follows. For every member $m \in [m^\ast]$, one has that $p_m(t) \rightarrow p_{m^\ast}(t)$ as $t \rightarrow \infty$. Hence, following a similar construction as above, one can choose $t'_0$ such that for all $t \geq t'_0$ and $m \in [m^\ast]$, the inequalities
\[
    \Bigl| p_m(t) - p_{m^\ast}(t) \Bigr| < \varepsilon'
\]
hold simultaneously for the precision $\varepsilon' > 0$. Applying this to the first sum yields the bounds
\[
    \sum_{m \in [m^\ast]} \bigl( p_{m^\ast}(t) - \varepsilon' \bigr) w_m(t)
    \leq \sum_{m \in [m^\ast]} p_m(t) w_m(t)
    \leq \sum_{m \in [m^\ast]} \bigl( p_{m^\ast}(t) + \varepsilon' \bigr) w_m(t).
\]
Here $\bigl( p_{m^\ast}(t) \pm \varepsilon' \bigr)$ are multiplicative constants that can be placed in front of the sum. Note that
\[
    1 \geq \sum_{m \in [m^\ast]} w_m(t)
    = 1 - \sum_{m \notin [m^\ast]} w_m(t)
    > 1 - \varepsilon.
\]
Use of the above inequalities allows simplifying the lower and upper bounds respectively:
\begin{equation}\label{eq:proof-3-c}
\begin{aligned}
    \bigl( p_{m^\ast}(t) - \varepsilon' \bigr)
    & \sum_{m \in [m^\ast]} w_m(t)
    > p_{m^\ast}(t) (1-\varepsilon') - \varepsilon'
    \geq p_{m^\ast}(t) - 2\varepsilon',
    \\ \bigl( p_{m^\ast}(t) + \varepsilon' \bigr)
    & \sum_{m \in [m^\ast]} w_m(t)
    \leq p_{m^\ast}(t) + \varepsilon'
    < p_{m^\ast}(t) + 2\varepsilon'.
\end{aligned}
\end{equation}

Combining the inequalities~(\ref{eq:proof-3-b}) and~(\ref{eq:proof-3-c}) in~(\ref{eq:proof-3-a}) yields the final result:
\[
    \Bigl| P(a_t|\g{\hat{a}o}_{<t}) - p_{m^\ast}(t) \Bigr| < 3\varepsilon' =
    \varepsilon,
\]
which holds with probability $\geq 1-\delta$ for arbitrary $\delta > 0$ related to $\delta'$ as $\delta' = 1 - \sqrt[M]{1-\delta}$ and arbitrary precision~$\varepsilon$.
\end{proof}

\subsection{Gibbs Sampling Implementation for MDP agent}
\label{subsec:gibbs}

Inserting the likelihood given in Equation~(\ref{eq:likelihood-reward}) into Equation~(\ref{eq:bayesian-control-rule}) of the Bayesian control rule, one obtains
the following expression for the posterior

\begin{eqnarray}
    \label{eq:posterior}
    P(m|\hat{a}_{\leq t}, o_{\leq t})
    &=& \frac{ P(x'|m, x, a) P(r|m, x, a, x')
            P(m|\hat{a}_{<t}, o_{<t}) }
    { \int_{\tilde{M'}} P(x'|m', x, a) P(r|m', x, a, x')
            P(m'|\hat{a}_{<t}, o_{<t}) \, dm' }
    \nonumber \\
    &=& \frac{ P(r|m, x, a, x')
            P(m|\hat{a}_{<t}, o_{<t}) }
    { \int_{\tilde{M'}} P(r|m', x, a, x')
            P(m'|\hat{a}_{<t}, o_{<t}) \, dm' },
\end{eqnarray}
where we have replaced the sum by an integration over $m'$, the
finite-dimensional real space containing only the average reward and the
Q-values of the observed states, and where we have simplified the term
$P(x'|m, x, a)$ because it is constant for all $m' \in \tilde{M'}$.

By inspection of Equation~(\ref{eq:posterior}), one sees that $m$ encodes
a set of independent normal distributions over the immediate reward having
means $\xi_m(x, a, x')$ indexed by triples $(x, a, x') \in
\fs{X}\times\fs{A}\times\fs{X}$. In other words, given $(x, a, x')$, the
rewards are drawn from a normal distribution with unknown mean $\xi_m(x,a,x')$
and known variance $\sigma^2$. The sufficient statistics are given by
$n(x,a,x')$, the number of times that the transition $x \rightarrow x'$ under
action $a$, and $\bar{r}(x, a, x')$, the mean of the rewards obtained in the
same transition. The conjugate prior distribution is well known and given by a
normal distribution with hyperparameters $\mu_0$ and $\lambda_0$:
\begin{equation}\label{eq:conjugate-prior}
    P(\xi_m(x,a,x')) = \normal(\mu_0, 1/\lambda_0)
    = \sqrt{\frac{\lambda_0}{2\pi}} \exp\Bigl\{ -\tfrac{\lambda_0}{2}
        \bigl( \xi_m(x,a,x')-\mu_0 \bigr)^2 \Bigr\}.
\end{equation}
The posterior distribution is given by
\[
    P(\xi_m(x,a,x')|\hat{a}_{\leq t}, o_{\leq t})
    = \normal(\mu(x,a,x'), 1/\lambda(x,a,x'))
\]
where the posterior hyperparameters are computed as
\begin{equation}
\begin{aligned}
    \mu(x,a,x') &= \frac{\lambda_0 \, \mu_0 + p \, n(x,a,x') \, \bar{r}(x,a,x')}
                {\lambda_0 + p \, n(x,a,x')}
    \\
    \lambda(x,a,x') &= \lambda_0 + p \, n(x,a,x').
\end{aligned}
\end{equation}
Finally, the conjugate distribution of the parameter vector $m$ is simply
the product
\begin{eqnarray}\label{eq:full-posterior}
    P(m|\hat{a}_{\leq t},o_{\leq t})
    &=& \prod_{x,a,x'} P(\xi_m(x,a,x')|\hat{a}_{\leq t}, o_{\leq t})
    \\ &\propto& \exp\Bigl\{
        -\frac{1}{2} \sum_{x,a,x'} \lambda(x,a,x')
        \bigl( \xi_m(x,a,x')-\mu(x,a,x') \bigr)^2
        \Bigr\}
\end{eqnarray}
because the $\xi_m(x,a,x')$ are independent but at the same time functions of
$m$. Thus, the MDP agent is
fully specified by the action probabilities in
Equation~(\ref{eq:prob-action}), the likelihood model in
Equation~(\ref{eq:likelihood-reward}), and the
prior distribution~(\ref{eq:conjugate-prior}).

Inference can be carried out by sampling $m$ from the posterior
distribution in Equation~(\ref{eq:full-posterior}). The actions issued by
the agent are by-products of the inference process. Here we derive an approximate
Gibbs sampler for $m$. We introduce the following symbols:
$m^{-\rho}$ and $m^{-Q(x,a)}$ stand for the parameter set removing
$\rho$ and $Q(x,a)$ respectively; $\mu$ and $\lambda$ are matrices collecting
the values of the posterior hyperparameters $\mu(x,a,x')$ and $\lambda(x,a,x')$
respectively; and $M(x) \define \max_{a} Q(x,a)$ is a shorthand.

Substituting $\xi_m(x,a,x')$ in Equation~(\ref{eq:full-posterior}) by its
definition (see Section~\ref{subsec:MDP}) and conditioning on the
Q-values, we obtain the conditional distribution of $\rho$:
\begin{equation}\label{eq:sampling-rho}
    P(\rho|m^{-\rho},\mu,\lambda)
    = \normal(\bar{\rho},1/S)
\end{equation}
where
\begin{gather*}
    \bar{\rho} = \frac{1}{S}\sum_{x,a,x'}
        \lambda(x,a,x') (\mu(x,a,x') - Q(x,a) + M(x')),
    \\ S = \sum_{x,a,x'} \lambda(x,a,x').
\end{gather*}

The conditional distribution over the Q-values is more difficult to obtain,
because each $Q(x,a)$ enters the posterior distribution both linearly and
non-linearly through $\mu$. However, if we fix $Q(x,a)$ within the $\max$
operations, which amounts to treating each $M(x)$ as a constant within a single
Gibbs step, then the conditional distribution can be approximated by
\begin{equation}\label{eq:sampling-q}
    P(Q(x,a)|m^{-Q(x,a)}, \lambda, \mu)
        \approx \normal\Bigl( \bar{Q}(x,a), 1/S(x,a) \Bigr)
\end{equation}
where
\begin{gather*}
    \bar{Q}(x,a) = \frac{1}{S(x,a)}\sum_{x'} \lambda(x,a,x')
        (\mu(x,a,x') - \rho + M(x')),
    \\ S(x,a) = \sum_{x'} \lambda(x,a,x').
\end{gather*}
We expect this approximation to hold because the resulting update rule
constitutes a contraction operation that forms the basis of most stochastic
approximation algorithms \citep{Mahadevan1996}. As a result, the Gibbs sampler
draws all the values from normal distributions. In each cycle of the adaptive
controller, one can carry out several Gibbs sweeps to obtain a sample of
$m$ to improve the mixing of the Markov chain. However, our experimental
results have shown that a \emph{single Gibbs sweep per state transition}
performs reasonably well.
Once a new parameter vector $m$ is drawn, the Bayesian control rule proceeds by taking the
optimal action given by Equation~(\ref{eq:prob-action}). Note that only the $\mu$
and $\lambda$ entries of the transitions that have occurred need to be
represented explicitly; similarly, only the Q-values of visited states need to
be represented explicitly.

\vskip 0.2in
%GATHER{bibliography.bib}
\bibliography{bibliography}

\end{document}